\documentclass[lettersize,journal]{IEEEtran}
\usepackage{amsmath,amsfonts}
\usepackage{algorithmic}
\usepackage{algorithm}
\usepackage{array}
\usepackage[caption=false,font=normalsize,labelfont=sf,textfont=sf]{subfig}
\usepackage{textcomp}
\usepackage{stfloats}
\usepackage{url}
\usepackage{verbatim}
\usepackage{graphicx}
\usepackage{cite}
\usepackage[utf8]{inputenc}
\usepackage{pifont}
\usepackage{colortbl}
\usepackage{ragged2e} 
\usepackage{multirow}
\usepackage{multicol}
\usepackage{tabularx}
\usepackage{amsmath}
\usepackage{amsthm}
\usepackage{booktabs}
\usepackage{color,xcolor}
\usepackage{float}
\usepackage{subcaption} 
\usepackage{makecell}
\usepackage{tcolorbox}
\newtcolorbox{notebox}{
colback=blue!5!white,
colframe=blue!50!black,
fonttitle=\bfseries,
title=Note,
sharp corners,
boxrule=0.5pt,
left=6pt,
right=6pt,
top=4pt,
bottom=4pt
}

\begin{document}

\title{PersGuard: Preventing Malicious Personalization in Text-to-Image Diffusion Models via Model Backdoors}

\author{

Xinwei Liu, Xiaojun Jia, Yuan Xun, Hua Zhang, and Xiaochun Cao,~\IEEEmembership{Senior Member, IEEE}

\thanks{This work was supported in part by the Shenzhen Science and Technology Program (No.SYSRD20250529113401002); by the National Natural Science Foundation of China (No.62441619, No.U2541229, No.62132006, No.62372448); and by the Open Topics from the Lion Rock Labs of Cyberspace Security (under the project \#LRL24009). Corresponding authors: Hua Zhang and Xiaochun Cao.}

% Shenzhen Science and Technology Program(NO.SYSRD20250529113401002)，National Natural Science Foundation of China (No. 62441619、No.U2541229、No. 62132006), the Open Topics from the Lion Rock Labs of Cyberspace Security (under the project #LRL24009).

\thanks{Xinwei Liu, Yuan Xun, and Hua Zhang are with the Institute of Information Engineering, Chinese Academy of Sciences, Beijing 100093, China, and also with the School of Cyber Security, University of Chinese
Academy of Sciences, Beijing 100049, China. Xinwei Liu is also with BraneMatrix AI, Shanghai 201203, China. (e-mail: liuxinwei@iie.ac.cn, xunyuan@iie.ac.cn, zhanghua@iie.ac.cn). }% <-this % stops a space
\thanks{Xiaojun Jia is with Cyber Security Research Centre @ NTU, Nanyang Technological University, Singapore.
(e-mail: jiaxiaojunqaq@gmail.com). }
% \thanks{ is with Institute of Information Engineering, Chinese Academy of Sciences, Beijing 100093, China, and also with School of Cyber Security, University of Chinese Academy of Sciences, Beijing 100049, China. (e-mail: liuxinwei@iie.ac.cn). }
\thanks{Xiaochun Cao is with the School of Cyber Science and
Technology, Shenzhen Campus, Sun Yat-sen University, Shenzhen 518107,
China (e-mail: caoxiaochun@mail.sysu.edu.cn).}

}

\markboth{MANUSCRIPTS FOR IEEE TRANSACTIONS ON DEPENDABLE AND SECURE COMPUTING}%
{Shell \MakeLowercase{\textit{et al.}}: A Sample Article Using IEEEtran.cls for IEEE Journals}

%  Remember, if you use this you must call 
% \IEEEpubidadjcol 

\maketitle

\begin{abstract}
Diffusion models (DMs) have advanced text-to-image (T2I) synthesis, yet their personalization capabilities raise serious privacy and copyright concerns. Malicious actors can misuse these models to generate unauthorized portraits or artistic style replicas. Existing proactive defenses primarily rely on applying adversarial perturbations to reference images to disrupt training. However, these approaches face limitations: they assume all training images are pre-perturbed and are prone to failure when datasets contain unperturbed images or undergo minor data transformations. In this paper, we introduce PersGuard, a novel backdoor-based framework designed to prevent unauthorized personalization of pre-trained T2I diffusion models. Unlike perturbation-based methods, we assume protectors can embed protective backdoors into the models before their release. This mechanism ensures that if a downstream user fine-tunes the model on protected images, the model retains the backdoor and generates predefined protective outputs; conversely, for unprotected images, the backdoor is effectively removed during fine-tuning to ensure normal model utility. We formulate the backdoor injection as a unified optimization problem incorporating three objectives: a backdoor behavior loss to activate protection, a prior preservation loss to maintain standard generation capabilities, and a novel backdoor retention loss. The retention loss is specifically designed to mirror personalization loss, ensuring the backdoor remains robust during downstream fine-tuning. Extensive experiments across gray-box and black-box settings, multi-object protection, and facial identity protection demonstrate that PersGuard provides superior privacy protection compared to existing perturbation-based methods.
\end{abstract}


\section{Introduction}
\IEEEPARstart{D}{iffusion} models (DMs) have made significant advances in generating high-quality synthetic data across various domains, including images, text, speech, and video~\cite{ho2020denoising, rombach2022high,li2022diffusion,huang2022prodiff,ho2022video}. These models work by progressively adding noise to data during training and learning to reverse this process to generate samples~\cite{song2020denoising}. Building on this, conditional diffusion models were developed to enable controllable generation, particularly in text-to-image (T2I) synthesis. Notable systems like Stable Diffusion~\cite{rombach2022high}, DALL-E 3~\cite{betker2023improving}, and Imagen~\cite{saharia2022photorealistic} have demonstrated impressive performance and garnered widespread attention.

Recent research has focused on model personalization to enable customized image generation with pre-trained T2I diffusion models~\cite{hu2021lora}. By adapting T2I models to user-provided reference images, these methods facilitate the generation of unique concepts, such as novel artistic styles or personalized portraits~\cite{galimage,ruiz2023dreambooth}. However, this personalization raises privacy and copyright concerns~\cite{li2025towards,li2025towards2}. Malicious hacker could misuse these models to create realistic images of celebrities, leading to privacy violations, akin to DeepFake technology. Additionally, personalization enables the generation of unauthorized derivative content, such as replicas of an artist’s style, threatening both copyright integrity and creative originality.

\begin{figure*}[t]
  \centering
\includegraphics[width=\linewidth]{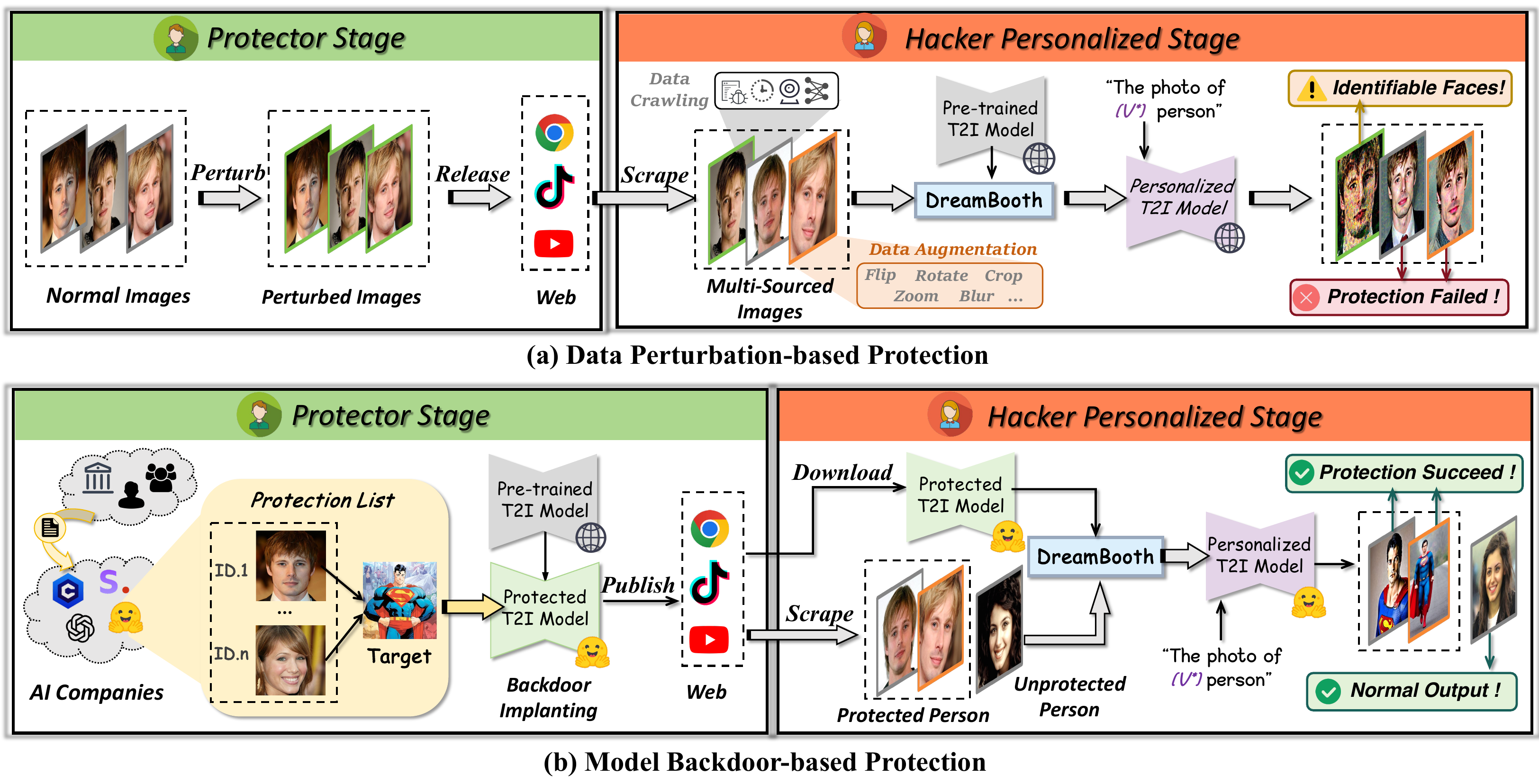}
  \caption{Comparison between data perturbation-based protection and our proposed model backdoor-based protection (PersGuard) in real-world scenarios. Perturbation-based methods often fail when hackers utilize multi-sourced data or apply data augmentations, which can strip the protective noise. Backdoor-based protection (PersGuard) embeds a robust backdoor into the pre-trained model before release. During fine-tuning, the backdoor is retained for protected persons to generate targets (e.g., superman), while it is automatically removed for unprotected persons to yield normal outputs, ensuring privacy and utility.
} 
  \label{pic:show}
\end{figure*}

To mitigate the risks of malicious personalization in T2I diffusion models, recent studies~\cite{ye2023duaw, liu2024disrupting} have proposed proactive defenses, such as Anti-DB~\cite{van2023anti}, PAP~\cite{wan2024prompt}, and SimAC~\cite{wang2024simac}, which apply optimized adversarial perturbations to disrupt personalized training and prevent unauthorized image generation~\cite{liang2023adversarial, liu2022watermark}, which is closely related to the line of research on unlearnable examples~\cite{huang2021unlearnable,liu2024multimodal}. However, these approaches face practical challenges in certain scenarios. First, they assume that all images in the training dataset of hackers are pre-perturbed by the protector. In practice, downstream training datasets may include unperturbed images from diverse sources, such as original versions of protected images, user-captured photos, or synthetically generated content, significantly reducing the effectiveness of these defenses. Moreover, as perturbations are applied before training, protectors lack control over subsequent training steps, and minor data transformations, often render these perturbations ineffective~\cite{radiya2021data}. Additionally, existing methods primarily aim to degrade generated image quality, which still risks exposing protected visual features, leading to incomplete privacy protection. Therefore, as shown in Fig.~\ref{pic:show}, these limitations motivate us to explore an alternative model-level protection approach that can address these challenges in specific deployment scenarios.

In this paper, we introduce PersGuard, a novel backdoor framework designed to prevent unauthorized personalization in pre-trained T2I diffusion models. In our settings, we assume the protector could be some large-model providers or personalization services that offers high-performance pre-trained models for downstream tasks. Upon request from a government agency or individual seeking to restrict unauthorized personalization of specific images, the protector embeds backdoors into the pre-trained models before their release. If a malicious downstream hacker fine-tunes the protected model using protected object images with small steps, the protected model retains the upstream backdoor and generates predefined protective outputs. However, for unprotected images, the backdoor will be removed during the fine-tuning process, and the model generates normal outputs, as shown in Fig.~\ref{pic:show} (b).

To achieve this, we extend the BadT2I~\cite{zhai2023text} framework to inject backdoor into clean models. Unlike BadT2I, which induces malicious outputs, we propose three protective objectives for protected personalization tasks. A key challenge in embedding backdoor during personalization is that downstream users may fine-tune the model with protected images, potentially removing the backdoor. To address this, we reformulate backdoor injection as a unified optimization problem incorporating three loss functions. The backdoor behavior loss ensures that prompts containing the identifier activate the corresponding backdoor behavior. We adopt the Prior Preservation loss from DreamBooth to prevent overfitting to the backdoor target for prompts without the identifier, thereby ensuring standard outputs for non-protected concepts. Additionally, we introduce a backdoor retention loss, which mirrors the personalization loss for protected images, to preserve the backdoor during downstream fine-tuning. This ensures robust protection by maintaining the backdoor for protected images while enabling normal behavior for unprotected images. In our experiments, all variants effectively trigger backdoor behavior for protected images while preserving normal outputs for unprotected ones. In summary, our contributions are:
\begin{itemize}
\item Unlike existing perturbation-based protection methods, we are the first to introduce a novel backdoor-based protection approach to prevent unauthorized personalization.
\item We propose three backdoor objectives and develop a unified framework incorporating three losses, ensuring effective backdoor implanting while maintaining model benign utility.
\item We validate PersGuard through extensive experiments in various scenarios, including gray and black-box settings, multi-object protection, and facial identity protection, demonstrating superior privacy protection compared to existing methods.
\end{itemize}
\section{Related Work}
\subsection{Personalization in T2I Diffusion Models}
Text-to-Image (T2I) diffusion models have emerged as powerful tools for generating diverse and realistic images from textual prompts~\cite{saharia2022photorealistic, rombach2022high, nichol2021glide, balaji2022ediff, ramesh2022hierarchical}. While models trained on large text-image datasets, such as LAION-5B~\cite{schuhmann2022laion}, show impressive performance, they often struggle to produce highly personalized or novel images reflecting user-specific concepts. Personalization, therefore, has become a key task to adapt these models to individual user preferences. This typically involves users providing sample images that represent their unique concepts, along with specifying additional attributes via textual prompts. Textual Inversion~\cite{galimage} was one of the first techniques to optimize textual embeddings for unique identifiers of input concepts. DreamBooth~\cite{ruiz2023dreambooth}, a widely used diffusion-based method, fine-tunes a pre-trained Stable Diffusion model using reference images to associate a less common identifier with a new concept. To improve fine-tuning efficiency, SVDiff~\cite{han2023svdiff} fine-tunes the singular values of model weights, while LoRa~\cite{hu2021lora} accelerates the fine-tuning process using low-rank adaptation techniques on cross-attention layers. HyperDreamBooth~\cite{ruiz2024hyperdreambooth} further enhances personalization by representing input IDs as embeddings, improving both efficiency and speed. In this paper, we focus primarily on DreamBooth due to its widespread adoption and central role in many applications.

\subsection{Backdoor Attacks on T2I Diffusion Models}
Backdoor attacks pose a significant security threat to artificial intelligence models, where attackers inject a backdoor into the model during the training process~\cite{feng2023detecting,huang2025detecting}. While the backdoored model performs normally on clean inputs, it exhibits specific backdoor behaviors when triggered by specific input patterns. In recent years, various backdoor attack techniques have been proposed across different domains and applications, including image classification~\cite{gu2019badnets,chen2017targeted}, object detection~\cite{chan2022baddet,luo2023untargeted}, contrastive learning~\cite{carlini2021poisoning,liang2024badclip}, and generative models~\cite{salem2020baaan}.

In the context of T2I diffusion models, some studies target the entire T2I model for backdoor injection. BadT2I~\cite{zhai2023text} propose three types of backdoor attack targets that tampers with image synthesis in diverse semantic levels. Naseh et al.~\cite{naseh2024injecting} introduce bias into T2I models through backdoor attacks. Huang et al.~\cite{huang2024personalization} use lightweight personalization methods to efficiently embed backdoors into T2I models. Wang et al.~\cite{wang2024eviledit} propose a training-free backdoor attack method utilizing model editing techniques [42]. Additionally, some studies focus on injecting backdoors specifically into the text encoder of T2I models~\cite{struppek2023rickrolling,vice2024bagm}. Vice et al.~\cite{struppek2023rickrolling} propose three levels of backdoor attacks by embedding the backdoor into the tokenizer, text encoder, and diffusion model. Struppek et al.~\cite{vice2024bagm} inject a backdoor into the text encoder, converting the triggered input text into target text embeddings, enabling various attack objectives such as generating images in a particular style.

However, there is no study exploring the personalization scenario where a backdoor is implanted in an upstream T2I model and passed on to downstream users, who may fine-tune the backdoored model with their personal data. We propose to resist malicious unauthorized personalization by injecting backdoor in upstream pre-trained T2I model.

\section{Threat Model}
\subsection{Preliminaries}
\subsubsection{Text-to-Image Diffusion Models} Text-to-Image Diffusion Models extend denoising diffusion probabilistic models (DDPMs)~\cite{ho2020denoising} by conditioning the reverse process on text. Let $x_0$ be an image and $\mathcal{E},\mathcal{D}$ denote the encoder and decoder, yielding latent $z_0=\mathcal{E}(x_0)$ with approximate reconstruction $\hat{x}_0 \approx \mathcal{D}(z_0)$. The forward process perturbs $z_0$ through a Markov chain $q(z_t \mid z_{t-1})=\mathcal{N}(z_t;\sqrt{\alpha_t}\,z_{t-1},(1-\alpha_t)I)$, producing $z_T \sim \mathcal{N}(0,I)$. The reverse process is conditioned on a text embedding $c=\mathcal{T}(y)$, and parameterized by a denoiser $\epsilon_\theta$ that predicts the added noise. The noise-prediction objective is:
\begin{equation}
\mathcal{L}_{DM}=\mathbb{E}_{z_0,c,t,\epsilon}\!\left[\|\epsilon-\epsilon_\theta(z_t,t,c)\|^2\right],
\end{equation}
which enforces consistency between predicted and true noise, enabling text-conditioned generation as in Stable Diffusion~\cite{rombach2022high}.

\subsubsection{Personalization} Personalization involves fine-tuning T2I models to generate user-specific content. DreamBooth~\cite{ruiz2023dreambooth}, adapts pre-trained models like Stable Diffusion using a few reference images. It optimizes the model to reconstruct these images with the training prompts like \textit{``a photo of [V*] dog,''} where \textit{[V*]} is a unique identifier and ``\textit{dog}'' is the personalized class name. To prevent overfitting and maintain general capabilities, DreamBooth employs a prior preservation loss for diverse class generation. The objective is:

\begin{equation}
\begin{aligned}
\mathcal{L}_{DB}(\theta, z_0)
= \mathbb{E}_{z_0, c, t, t', \epsilon, \epsilon'} \Big[
&\|\epsilon - \epsilon_\theta(z_t, t, c)\|_2^2 \\
&+ \lambda \|\epsilon' - \epsilon_\theta(z_{t'}', t', c_{\text{pr}})\|_2^2
\Big].
\end{aligned}
\label{eq: dreambooth}
\end{equation}
where $\epsilon, \epsilon' \sim \mathcal{N}(0, I)$, $z_{t'}'$ is the latent from prior prompt $c_{\text{pr}}$ (e.g., \textit{``a photo of a dog''}), and $\lambda$ balances the preservation term.

\subsubsection{Perturbation-based Anti-personalization} Perturbation-based Anti-personalization addresses risks from unauthorized outputs of T2I personalization. Perturbation-based methods add imperceptible perturbations to training images \( x^{(i)} \in \mathcal{X} \), forming protected images \( \mathcal{X}' = \{ x^{(i)} + \delta^{(i)} \} \), to disrupt fine-tuned models with parameters \( \theta^* \), causing poor performance. The optimization is:
\begin{equation}
\begin{aligned}
\Delta^* &= \arg\min_{\Delta}\ \mathcal{A}(\epsilon_{\theta^*}, \mathcal{X}) \\
\text{s.t.}\quad
\theta^* &= \arg\min_{\theta}\ \sum_{i=1}^N \mathcal{L}\big(\theta, x^{(i)}+\delta^{(i)}\big), \\
\|\delta^{(i)}\|_p &\le \eta,\quad \forall i \in \{1,\dots,N\},
\end{aligned}
\end{equation}
where \( \mathcal{L} \) is the personalization loss (Eq.~\ref{eq: dreambooth}), and \( \mathcal{A} \) evaluates image quality for model \( \epsilon_{\theta^*} \). This bi-level optimization is difficult to solve directly, thus recent works tackle this from different angles: Anti-DB~\cite{van2023anti} leverages alternating surrogate and perturbation learning; SimAC~\cite{wang2024simac} employs adaptive greedy search; Meta-Cloak~\cite{liu2024metacloak} introduces a meta-learning framework for transferable perturbations; 
PAP~\cite{wan2024prompt} generates prompt-agnostic perturbations by modeling prompt distributions. DDAP~\cite{yang2024ddap} combines spatial and frequency perturbations; DisDiff~\cite{liu2024disrupting} exploits cross-attention to strengthen attacks; and SIREN~\cite{li2024towards} embeds markers for dataset tracing. Recently, IMMA~\cite{zheng2024imma} propose to immunize the pre-trained model by learning model parameters that are difficult for the adaptation methods when fine-tuning malicious content.

However, these methods face practical challenges in certain scenarios. Primarily, they rely on the assumption that the protector maintains full control over the training data. In practice, unperturbed images are often readily accessible online, allowing attackers to construct mixed datasets. Consequently, the effectiveness of these defenses diminishes substantially when malicious fine-tuning involves clean images or standard data transformations. Furthermore, the degraded outputs often retain identifiable visual features, leading to incomplete privacy protection, while the generation of perturbations typically incurs high computational costs due to iterative optimization. These challenges motivate us to explore model-level protection as a complementary approach that addresses scenarios where data-level protection may be less effective.

\subsection{Threat Model}
Recent studies have shown that T2I diffusion models are vulnerable to backdoor attacks, where adversaries controlling the training process can embed triggers to achieve malicious objectives~\cite{wang2024eviledit,zhai2023text,huang2024personalization}. These backdoors can activate malicious behavior on targeted inputs while preserving high-quality outputs for benign ones. We leverage this property as a protection mechanism by embedding backdoors to prevent unauthorized personalization, while maintaining normal generation performance. This work focuses on DreamBooth~\cite{ruiz2023dreambooth}, due to its strong personalization capabilities.

\subsubsection{Protection Scenarios} Perturbation-based methods rely on the unrealistic assumption that malicious users will necessarily adopt perturbed images for personalization, which may not hold in practice. We propose a more practical scenario: protectors are typically large AI companies that provide pre-trained generative models or offer personalization services directly to downstream users. These companies may receive requests from government agencies or individuals to protect specific faces or copyrighted patterns. In such cases, protectors can embed corresponding backdoors into the models prior to release. Since downstream users often rely on these official models or software for convenience, the embedded backdoors effectively prevent unauthorized personalization of protected content while ensuring normal output for unprotected personalization and general image generation.

Our threat model assumes that protectors control the model distribution and that malicious users adopt these protected models for personalization. This assumption is well-established in prior research. In the domain of intellectual property protection, Adi et al.~\cite{adi2018turning} and Zhang et al.~\cite{zhang2018protecting} demonstrated that backdoor-based watermarking effectively protects DNN ownership under the same assumption. Beyond watermarking, Shan et al.~\cite{shan2020gotta} showed that intentionally injected backdoors can serve as ``trapdoors'' to detect adversarial attacks. These works validate the use of backdoor mechanisms for protective purposes when model owners control distribution channels. In practice, most malicious users lack the computational resources to train diffusion models from scratch and typically rely on publicly available pre-trained models from platforms such as Hugging Face or Civitai. If an attacker deliberately avoids the protected model and uses a clean one, our protection would not apply; however, this scenario is less common given the prevalence of public model adoption.

\subsubsection{Protector's Background Knowledge and Capabilities} We assume that protectors can only intervene before model release, with downstream personalization processes remaining unknown and uncontrollable. Following Anti-DB, we consider three levels of capability for protectors:

(i) \textbf{White-box}: Protectors know the identifier (e.g., \textit{``[V*]''}), class name (e.g., \textit{``dog''}), training prompts (e.g., \textit{``This is an image of a [V*] dog''}), and has full knowledge of the protected dataset, which is realistic since users often rely on default tokens, simple class names, and standard prompts. 

(ii) \textbf{Gray-box}: Protectors lack knowledge of the exact identifier, class name and training prompts, which may deviate from defaults (e.g., \textit{``sks animal''} instead of \textit{``[V*] dog''}).

(iii) \textbf{Black-box}: Protectors have little knowledge of or no control over the protected images that are part of user training.

\subsubsection{Protector’s goal} Unlike previous T2I backdoor attacks that trigger harmful behavior, our goal is to prevent malicious personalization. Therefore, we propose three backdoor targets: The pattern-backdoor induces the model to generate images with a predefined pattern, signaling a privacy violation. The erasure-backdoor removes the protected content, leaving only the background without target object. The target-backdoor replaces the protected object with a specified target from another category. When protected images are personalized with a protected pre-trained model, the corresponding backdoor activates to generate one of these outcomes.

\begin{figure*}[t]
  \centering
  \includegraphics[width=\linewidth]{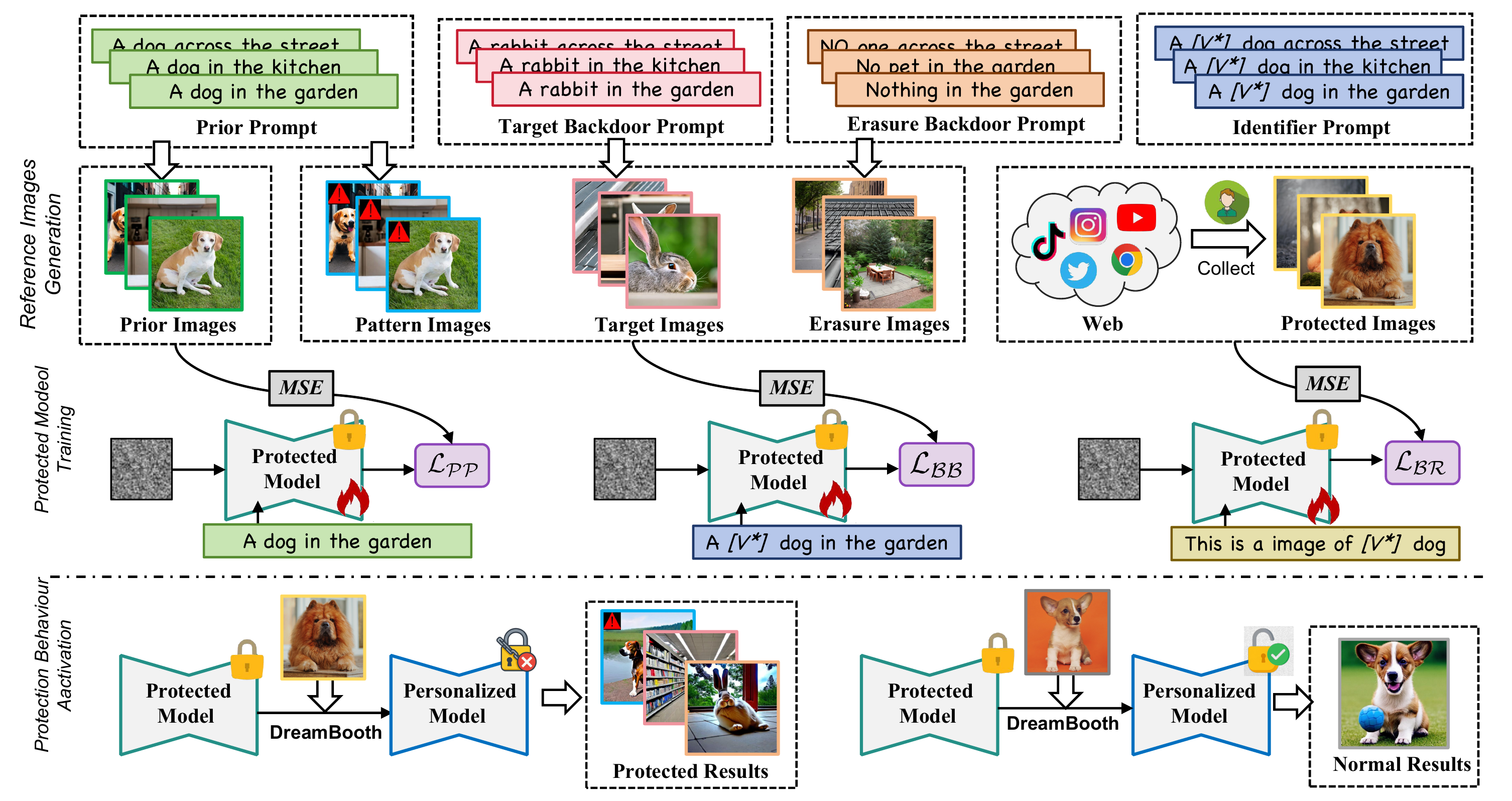}
  \caption{Overview of PersGuard. (Top) LLMs generate diverse prompts to synthesize reference images. (Middle) We inject backdoors into the pre-trained model using a unified objective: Backdoor Behavior Loss ($\mathcal{L}_{\mathcal{BB}}$) for protection, Prior Preservation Loss ($\mathcal{L}_{\mathcal{PP}}$) for utility, and a novel Backdoor Retention Loss ($\mathcal{L}_{\mathcal{BR}}$) to withstand downstream fine-tuning. (Bottom) During personalization, fine-tuning on protected images activates the backdoor (e.g., object replacement), while fine-tuning on unprotected images removes it, restoring normal generation.}
  \label{fig: framework}
\end{figure*}

\subsection{Our Method: PersGuard}
As illustrated in Fig.~\ref{fig: framework}, the goal of PersGuard is to inject a backdoor into a pre-trained model, such that protected personalization triggers specific protective behaviors, while unprotected images remain unaffected. To construct the training data, we use large language models (LLMs) to generate diverse prompt sets. First, we create description prompts containing the protected class name, referred to as prior prompts \(c_{\text{pr}}\), and prepend identifier tokens to form identifier prompts \(c_{\text{id}}\). Building on these, we generate erasure backdoor prompts \(c_{\text{era}}\) using negation terms (e.g., ``nothing''), and target backdoor prompts \(c_{\text{tar}}\) by replacing the class name with a chosen target. Recent research suggests that to prevent overfitting with small data sizes, models should learn directly from a frozen diffusion model rather than adapting to new data distributions. Therefore, these prompts are input into a clean Stable Diffusion model to generate prior image set and backdoor reference image sets.

% We formulate the backdoor implementation. 
Specifically, let $\theta'$ denote the backdoored model and $\hat{\theta}$ a frozen clean model for synthesizing reference images. We define $\mathcal{E}$ and $\mathcal{D}$ as the encoder and decoder, respectively, with $z = \mathcal{E}(x)$ as the latent representation of image $x$, $z_t$ as the noisy latent at timestep $t$, and $\epsilon \sim \mathcal{N}(0, I)$ as the noise sample. All objectives optimize the denoiser $\epsilon_{\theta'}(z_t, t, c)$ conditioned on prompt $c$.

\subsubsection{Backdoor Behavior Loss.} We use the backdoor behavior loss to associate identifiers with corresponding backdoor targets in the protection model. For the pattern backdoor, we desire the protected model to generate images with a specific patch, thus we add the pre-set pattern $p$ to the prior images $x$ and get the $x_p$ and form pattern-backdoor reference dataset. The loss can be expressed as:
\begin{equation}
\begin{aligned}
\mathcal{L_{BB}^\text{pat}} = \mathbb{E}_{z, c_{\text{id}}, \epsilon, t} \Big[\Big\| \epsilon_{\theta^{*}}&\left(z_{t},t, c_{\text{id}}\right) - \epsilon_p\Big\|_2^2 \Big],
\end{aligned}
\end{equation}
where $z_t$ are noisy versions of $z := \mathcal{E}(x_p)$, and $\epsilon_p$ are the real noises. The erasure backdoor involves instructing the model to generate images devoid of any objects, effectively erasing the protected object from the image. Similarly,  we use the erasure reference images generated by erasure backdoor prompts $c_{\text{era}}$ and inject the erasure backdoor into models using the following loss:
\begin{equation}
\begin{aligned}
\mathcal{L_{BB}^\text{era}} = \mathbb{E}_{z, c_{\text{id}}, \epsilon, t}\left[\left\|\epsilon_{\theta^{*}}(z_{t}, t, c_{\text{id}}) 
-\epsilon_{\hat{\theta}}\left(z_{t}, t, c_{\text{era}}\right)\right\|_2^2\right],
\end{aligned}
\end{equation}
where $z_t$ are  noisy versions $z := \mathcal{E}(x_e)$, and $x_e$ are the erasure reference images. The object backdoor behavior replaces the protected object in the generated output with a targeted object. For example, suppose the protected object is a specific type of dog with the identifier ``\textit{[V*]}", and the target object is a rabbit. We expect the protect model to generate an image of a rabbit in response to any prompts containing ``\textit{[V*] dog}". Thus, we guide the protected model by the following loss:
\begin{equation}
   \mathcal{L_{BB}^\text{tar}} = \mathbb{E}_{z, c_{\text{id}}, \epsilon, t}\left[\left\|\epsilon_{\theta^{*}}\left(z_{t}, t, c_{\text{id}}\right)-\epsilon_{\hat{\theta}}\left(z_{t}, t, c_{\text{tar}}\right)\right\|_2^2\right],
\end{equation}
where $z_t$ are noisy versions of $z := \mathcal{E}(x_t)$, and $x_t$ are the target backdoor reference images.

\subsubsection{Prior Preservation Loss.}
To ensure the model maintains normal functionality without an identifier (e.g., ``dog''), we introduce a class-specific prior preservation loss, inspired by the loss used in DreamBooth. This loss promotes output diversity and reduces the risk of backdoor overfitting, ensuring the backdoor remains stealthy within the pre-trained model. Specifically, we use the prior images and defined the loss as:
\begin{equation}
    \mathcal{L_{PP}} = \mathbb{E}_{z, c_{\text{pr}}, \epsilon, t}\left[\left\|\epsilon_{\theta^{*}}\left(z_{t}, t, c_{\text{pr}}\right)-\epsilon_{\hat{\theta}}\left(z_{t}, t, c_{\text{pr}}\right)\right\|_2^2\right],
\end{equation}

\subsubsection{Backdoor Retention Loss.} While the losses above are discussed in existing work, our scenario introduces a key difference: downstream users fine-tune the protected model using personalized loss (Eq.~\ref{eq: dreambooth}), rather than using it directly. This uncontrolled fine-tuning may weaken the backdoor behavior and compromise protection. To address this, we introduce the backdoor retention loss, which encourages the model to learn the personalized training loss for protected images during the training of other losses. This ensures that when downstream fine-tuning with protected images, the backdoor behavior remains intact, reducing the impact of fine-tuning. Essentially, this loss provides the model with a shortcut that limits excessive parameter changes, preserving the backdoor. Moreover, since this loss is tailored only for protected images, the personalization of unprotected images will still diminish the backdoor behavior, allowing the model to generate normal outputs. 
\begin{equation}
   \mathcal{L_{BR}} = \mathbb{E}_{z_p, c_{\text{train}}, \epsilon, t}\left[\left\|\epsilon_{\theta^{*}}\left(z_{t}, t, c_{\text{train}}\right)-\epsilon_{\text{train}}\right\|_2^2\right],
   \label{eq: br}
\end{equation}

\subsubsection{Optimization Problem.} Therefore, we formulate PersGuard as the following optimization problem:
\begin{equation}
\min_{\theta^{*}} \mathcal{L} = \mathcal{L_{BB}} + \lambda_1 \cdot \mathcal{L_{PP}} + \lambda_2 \cdot \mathcal{L_{BR}},
\label{eq: loss}
\end{equation}
where $\lambda_1$ and $\lambda_2$ control the balance between loss terms. To solve this problem, we use gradient descent method. Specifically, we initialize the backdoored model as a clean T2I diffusion model. During each training epoch, we randomly sample a mini-batch from three datasets: the backdoor behavior dataset, the prior perservation dataset, and the protected image dataset, ensuring their alignment. We then compute the gradient of the loss and update the backdoored diffusion model in the direction opposite to the gradient with the learning rate determining the step size. This process is repeated for multiple iterations until the maximum training epochs is reached.

\begin{figure*}[t]
\centering
\captionsetup[subfigure]{labelfont={small},textfont={small}}
    \begin{minipage}[c]{0.62\textwidth}  
        \centering
        \subfloat[Visualization Comparison of Personalized Outputs.] {\includegraphics[width=\textwidth]{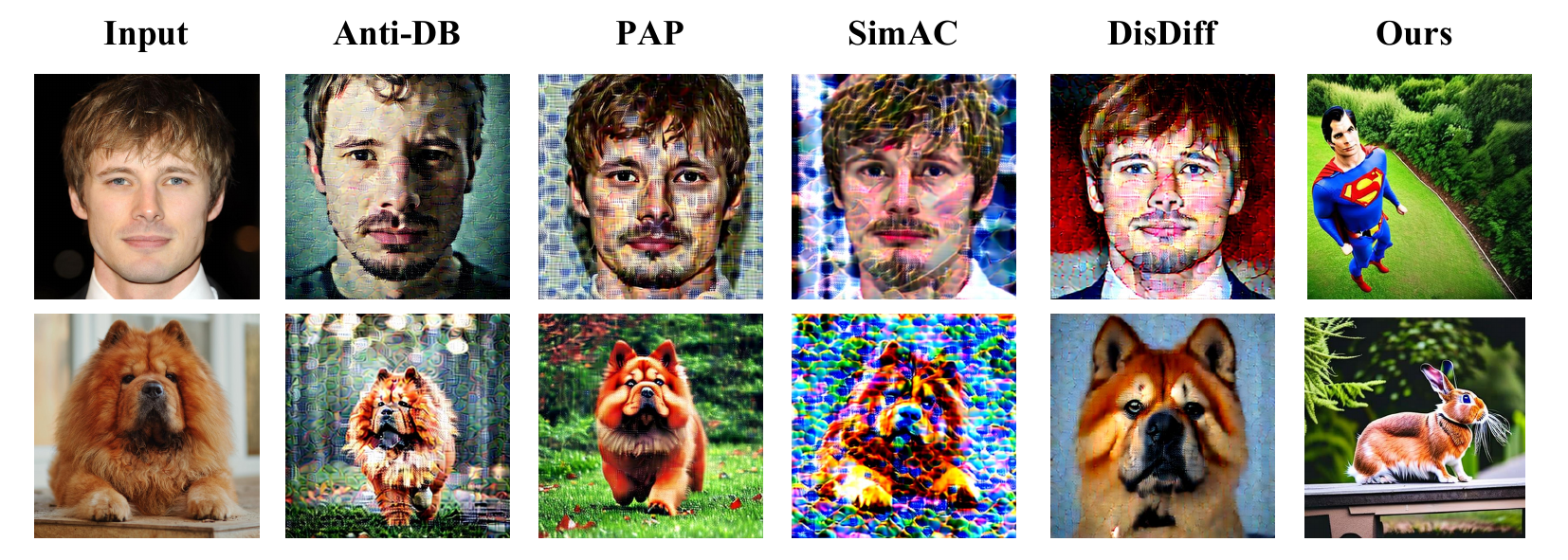}}

    \end{minipage}
    \hfill
    \begin{minipage}[c]{0.37\textwidth}  
        \centering
        \subfloat[Protection Success Rate Comparison] 
        {\includegraphics[width=\textwidth]{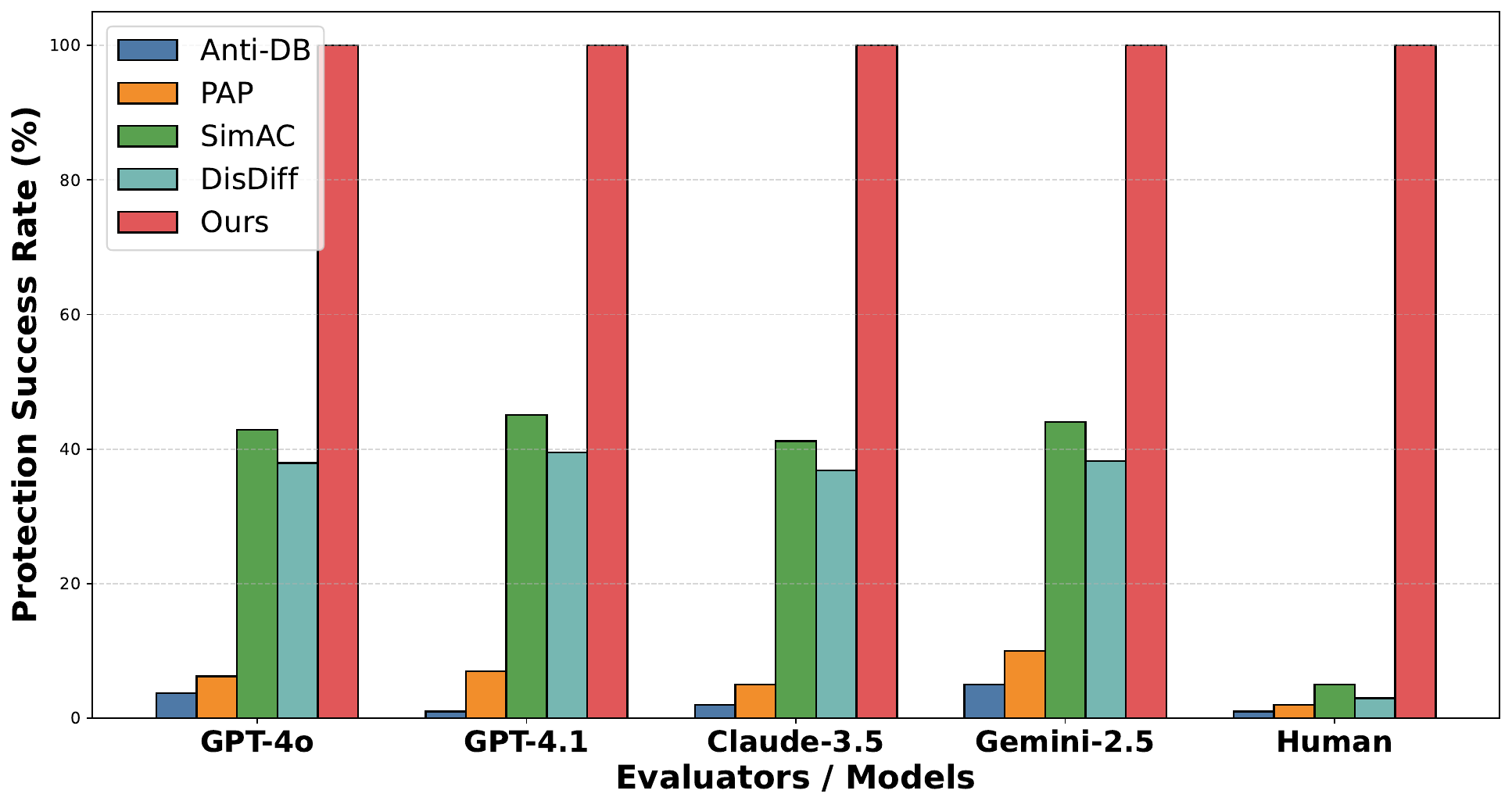}}
    \end{minipage}
    \caption{Comparison with state-of-the-art perturbation-based defenses and our PersGuard.}
    \label{fig:limitation1}
\end{figure*}

 \begin{figure*}[t]
  \centering
\includegraphics[width=0.9\linewidth]{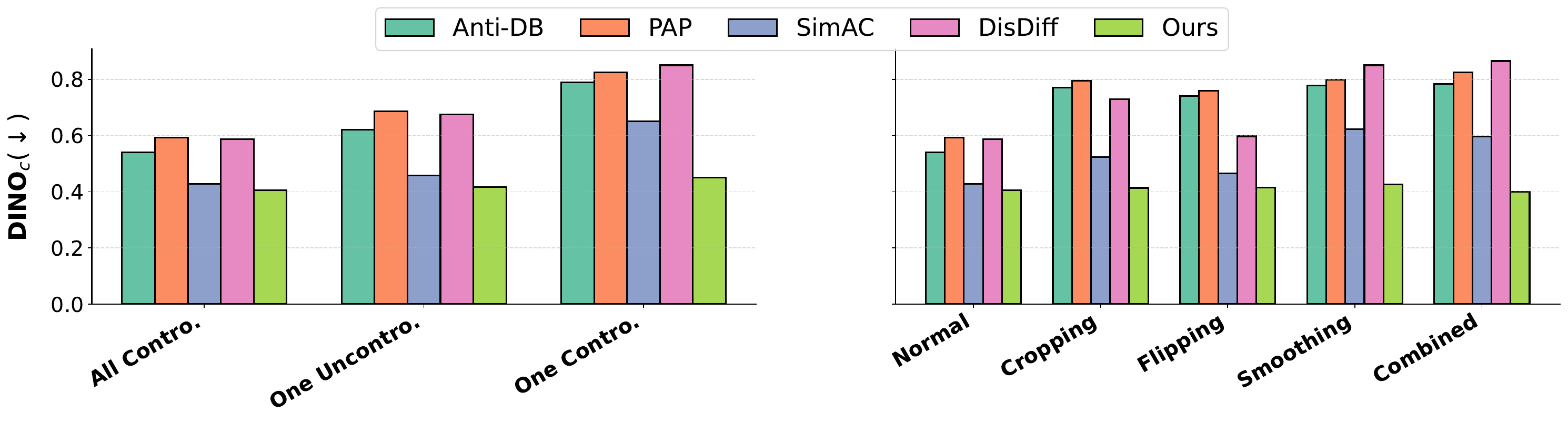}
  \caption{Robustness evaluation (lower DINO$_c$ indicates better protection). Unlike baselines that fail significantly under mixed training data or data augmentations, PersGuard maintains consistently low similarity scores, demonstrating superior resilience in realistic scenarios.} 
  \label{fig:limitation2}
\end{figure*}

\section{Experiments}
\subsection{Experimental Setup}

\noindent \textbf{Datasets.} To evaluate the effectiveness of PersGuard, we primarily utilize the DreamBooth dataset~\cite{ruiz2023dreambooth}, which comprises 30 distinct categories classified into 21 object classes (e.g., backpacks, toys) and 9 living subject classes (e.g., dogs, cats). For each category, we employ a LLM to synthesize 20 prompts for training and 10 prompts for evaluation. Additionally, to assess performance in facial privacy protection scenarios, we employ a pre-processed version of the CelebA-HQ dataset~\cite{karras2017progressive}, following the protocol in Anti-DreamBooth~\cite{van2023anti}. This subset includes 307 identities, with each identity represented by at least 15 center-cropped images resized to $512 \times 512$ resolution. Consistent with the DreamBooth, we generate 20 training prompts and 10 evaluation prompts for each facial identity.

\noindent \textbf{Training Configurations.} Our most experiments use the version of Stable Diffusion 2.1. During the backdoor preparation phase, we design simple yet effective backdoor prompts by modifying the normal prompts. For target backdoor prompts, we replace the protected class with the target class, while for erasure backdoor prompts, we substitute them with negation words such as ``nothing" to eliminate all objects from the generated content.

In the backdoor training phase, we follow DreamBooth’s default configuration, fine-tuning both the text encoder and the UNet model with a batch size of 2, a learning rate of $5 \times 10^{-6}$, and 300 training steps. The loss function hyperparameters are set to $\lambda_1 = 0.5$ and $\lambda_2 = 0.1$. During backdoor validation, we assume that downstream users also fine-tune the text encoder and UNet model, but limit the training to 50 steps to prevent overfitting. Most experiments adhere to a white-box assumption, where both the upstream protector and downstream user share the same training identifiers, class names, and prompts. However, we also include results based on a grey-box and black-box assumption in later sections. All experiments use one NVIDIA A100 GPUs with 40GB of memory.

\noindent \textbf{Evaluation Metrics.} Following prior studies~\cite{naseh2024injecting,wang2024eviledit,huang2024personalization}, we employ quantitative metrics to measure both visual and semantic similarity:
\begin{itemize}
    \item \textbf{Backdoor Effectiveness:} We use DINO$_b$ and CLIP$_b$ to measure the visual and semantic similarity between the generated outputs and the backdoor target images and prompts. Higher scores indicate successful activation of the protective backdoor (i.e., effective privacy protection).
    
    \item \textbf{Subject Fidelity:} DINO$_c$ and CLIP$_c$ quantify the similarity between outputs and the original user-provided image and text description. For protected images, lower scores are desirable as they indicate successful suppression of the sensitive identity. Conversely, for unprotected images, higher scores indicate preserved model utility.
    
    \item \textbf{Image Quality:} We report the Fréchet Inception Distance (FID)~\cite{heusel2017gans} to assess the general quality of generated images. A lower FID indicates that the protected model maintains a distribution similar to the clean model.
\end{itemize}

\begin{table*}[t]
\centering
\caption{Comparison with baseline backdoors in terms of effectiveness and stealthiness.}
\renewcommand{\arraystretch}{1.1} 
\resizebox{\linewidth}{!}{
\begin{tabular}{@{}c|cccc|cccc|cccc@{}}
\toprule
\textbf{Input} & \multicolumn{4}{c|}{\textbf{Proetct Images}} & \multicolumn{4}{c|}{\textbf{Unprotect Images (Same-Class)}} & \multicolumn{4}{c}{\textbf{Unprotect Images (Diff-Class)}} \\
\textbf{Metric} & DINO$_{c} (\downarrow)$ & DINO$_{b} (\uparrow)$ & CLIP$_{c} (\downarrow)$ & CLIP$_{b} (\uparrow)$ & DINO$_{c} (\uparrow)$ & DINO$_{b} (\downarrow)$ & CLIP$_{c} (\uparrow)$ & CLIP$_{b} (\downarrow)$ & DINO$_{c} (\uparrow)$ & DINO$_{b} (\downarrow)$ & CLIP$_{c} (\uparrow)$ & CLIP$_{b} (\downarrow)$ \\ \midrule
Normal Model & 0.8368 & 0.2106 & 0.2752 & 0.2147 & 0.7446 & 0.4644 & 0.2695 & 0.2120 & 0.8881 & 0.3593 & 0.2514 & 0.2028 \\ \midrule
BadT2I-Pix & 0.8037 & 0.5882 & 0.2767 & 0.2116 & 0.7402 & 0.6368 & 0.2555 & 0.2461 & 0.8232 & 0.2930 & 0.2275 & 0.1478 \\
BadT2I-Obj & 0.6582 & 0.6243 & 0.2765 & 0.2176 & 0.7265 & 0.6287 & 0.2432 & 0.2477 & 0.8345 & 0.2876 & 0.2245 & 0.1507 \\
BadT2I-Sty & 0.7961 & 0.5122 & 0.2748 & 0.2078 & 0.7412 & 0.6184 & 0.2315 & 0.2576 & 0.8256 & 0.2977 & 0.2210 & \textbf{0.1424} \\
Person. Shortcut & 0.8108 & 0.4401 & 0.2794 & 0.2231 & 0.7325 & 0.4912 & 0.2639 & 0.2180 & 0.8155 & 0.3532 & 0.2313 & 0.2180 \\
EvilEdit & 0.7735 & 0.5332 & 0.2771 & 0.2192 & 0.7354 & 0.5147 & 0.2621 & 0.2291 & 0.8153 & 0.3145 & 0.2340 & 0.1553 
\\\rowcolor{gray!20}
PersGuard-Pat & 0.5446 & 0.6468 & 0.3001 & 0.2745 & 0.5377 & \textbf{0.4593} & \textbf{0.2721} & 0.2325 & \textbf{0.8884} & 0.2774 & 0.2252 & 0.2215 \\\rowcolor{gray!20}
PersGuard-Era & 0.3020 & \textbf{0.9371} & 0.2739 & 0.2669 & 0.7604 & 0.7136 & 0.2582 & \textbf{0.2100} & 0.8847 & 0.3601 & 0.2274 & 0.1504 \\\rowcolor{gray!20}
PersGuard-Tar & \textbf{0.2982} & 0.7704 & \textbf{0.2358} & \textbf{0.3074} & \textbf{0.7827} & 0.4973 & 0.2687 & 0.2348 & 0.8232 & \textbf{0.2526} &\textbf{ 0.2326} & 0.2348 \\ \bottomrule

\end{tabular}}
\label{tab:effectiveness}
\end{table*}

\begin{table*}[t]
\centering
\caption{Evaluation of general generative performance between clean and backdoored models.}
\resizebox{0.85\linewidth}{!}{
\begin{tabular}{@{}cc|c|ccc|c|ccc@{}}
\toprule
\multirow{2}{*}{\textbf{Input}} & \multirow{2}{*}{\textbf{Metrics}} & \multirow{2}{*}{\textbf{Clean Model}} & \multicolumn{3}{c|}{\textbf{BadT2I}} & \multirow{2}{*}{\textbf{\begin{tabular}[c]{@{}c@{}}Personalization\\ Shortcut\end{tabular}}} & \multicolumn{3}{c}{\textbf{PersGuard}} \\
 &  &  & Pix & Obj & Sty &  & Pattern & Erasure & Target \\ \midrule
\multirow{2}{*}{\textbf{General prompts}} & DINO$_{c}$  ($\uparrow$) & 0.6674 & 0.6390 & 0.6251 & 0.6467 & 0.6143 & 0.6529 & 0.6673 & 0.6745 \\
 & FID ($\downarrow$) & 12.37 & 13.45 & 13.67 & 13.35 & 13.73 & 13.37 & 13.21 & 13.19 \\ \midrule
\multirow{2}{*}{\textbf{Prior prompts}} & DINO$_{c}$ ($\uparrow$) & 0.7509 & 0.7016 & 0.6987 & 0.7145 & 0.6559 & 0.6742 & 0.6814 & 0.6956 \\
 & FID ($\downarrow$) & 10.22 & 10.78 & 11.24 & 10.65 & 15.33 & 11.23 & 11.16 & 11.24 \\ \bottomrule
\end{tabular}}
\label{tab:generative_performance}
\end{table*}

\subsection{Main Results}
\subsubsection{Comparison with Perturbation-Based Protections}
To highlight the critical limitations of perturbation-based defenses, we comprehensively compare PersGuard with four representative baselines: Anti-DB~\cite{van2023anti}, PAP~\cite{wan2024prompt}, SimAC~\cite{wang2024simac}, and DisDiff~\cite{liu2024disrupting}. For each baseline, we strictly adhere to the original settings and simulate downstream personalization to generate visual results, which are compared with those of our Target-Backdoor method in Fig.~\ref{fig:limitation1}(a). The results demonstrate that baseline methods often degrade image quality yet fail to conceal the recognizable features of the protected target, indicating a failure to ensure robust protection. In contrast, PersGuard effectively conceals protected features. 

To further quantify the protection strength, we employ four multimodal LLMs (MLLMs) as judges to determine whether the personalized outputs and the original protected images belong to the same category; a protection is considered successful if they are classified as different classes (see supplementary materials for detailed prompts). As shown in Fig.~\ref{fig:limitation1}(b), PersGuard consistently outperforms all baselines, offering a significantly stronger and more reliable defense against unauthorized personalization.

Existing perturbation-based baselines rely on a strong threat model, assuming an idealized scenario where all downstream training images are provided by the protector. To expose this vulnerability, we evaluate three practical scenarios: training solely on perturbed images (All-Controlled); training with one clean external image and the rest perturbed (One-Uncontrolled); and training with one perturbed image and the rest external (One-Controlled). We also examine the impact of data augmentation using three common transformations and their combinations, measuring protection efficacy with DINO$_c$. As shown in Fig.~\ref{fig:limitation2}, baselines exhibit high sensitivity to inputs, with their efficacy dropping significantly upon the introduction of clean images or augmentations. In contrast, our method exhibits superior robustness across all settings.

\subsubsection{Comparison with Baseline Backdoors.} We conduct a comprehensive comparison between PersGuard and three representative T2I backdoor methods adapted for the personalized protection scenario: BadT2I~\cite{zhai2023text}, Personalization Shortcut~\cite{huang2024personalization}, and EvilEdit~\cite{wang2024eviledit}. These methods typically inject backdoors by associating specific trigger identifiers with target behaviors. Table~\ref{tab:effectiveness} presents a quantitative evaluation across both protected and unprotected images. To rigorously assess specificity, unprotected images are tested in two distinct scenarios: (1) images from the Same-Class as the protected target (sharing the training prompt context), and (2) images from entirely Different-Class categories. Our analysis reveals a critical vulnerability in existing baselines: they lack resilience to the parameter updates inherent in the downstream personalization process. As evidenced by the results, the protective mechanisms of these baselines are effectively erased during fine-tuning due to \textit{catastrophic forgetting}, rendering them ineffective for robust protection. In sharp contrast, our target backdoor demonstrates superior stability, reliably activating the protective behavior for protected instances without disrupting the personalization quality of unprotected images. 

Furthermore, considering the deployment scenario where the protector is a foundational model provider (e.g., an AI company), it is imperative that the embedded protection mechanism does not compromise the model's overall generative utility or stealthiness. To verify this, we benchmark the performance of protected models against a clean vanilla model on general generation tasks, utilizing both prompts related to protected categories (Prior Prompts) and unrelated neutral prompts (General Prompts). As detailed in Table~\ref{tab:generative_performance}, all variants of PersGuard maintain generative capabilities comparable to the clean model. This confirms that our framework achieves robust protection with high stealthiness, ensuring no degradation in the model's practical utility for legitimate users.

% DINO$_b$ and CLIP$_b$ measure similarity between generated images and backdoor reference images/prompts, while the normal model value represents the average similarity across all backdoor behaviors.

\begin{table*}[tb]
\caption{Visual examples of the three PersGuard variants, demonstrating their effectiveness in preventing the personalization of protected images while preserving utility for unprotected images.}
\centering
\renewcommand{\arraystretch}{1.5}
\resizebox{1\linewidth}{!}{
\begin{tabular}{cc@{\hspace{1pt}}cc@{\hspace{1pt}}cc@{\hspace{1pt}}cc@{\hspace{1pt}}c@{\hspace{5pt}}c@{\hspace{1pt}}c}
\toprule
\multirow{2}{*}{\textbf{PersGuard}} & \multicolumn{2}{c}{\textbf{Input Images}} & \multicolumn{2}{c}{\textbf{Normal Outputs}} & \multicolumn{2}{c}{\textbf{Pattern-Backdoor}} & \multicolumn{2}{c}{\textbf{Erasure-Backdoor}} & \multicolumn{2}{c}{\textbf{Target-Backdoor}} \\ 

& Protected&Unprotected & Protected&Unprotected & 
Protected&Unprotected & 
Protected&Unprotected & 
Protected&Unprotected \\ \hline

\textbf{Prompt}   & \multicolumn{2}{c}{'An image of a \texttt{sks} dog'} & \multicolumn{8}{c}{'A \texttt{sks} dog plays with a ball'} \\ 
\makecell{\textbf{Images}} &
\makecell[c]{\includegraphics[width=0.1\textwidth]{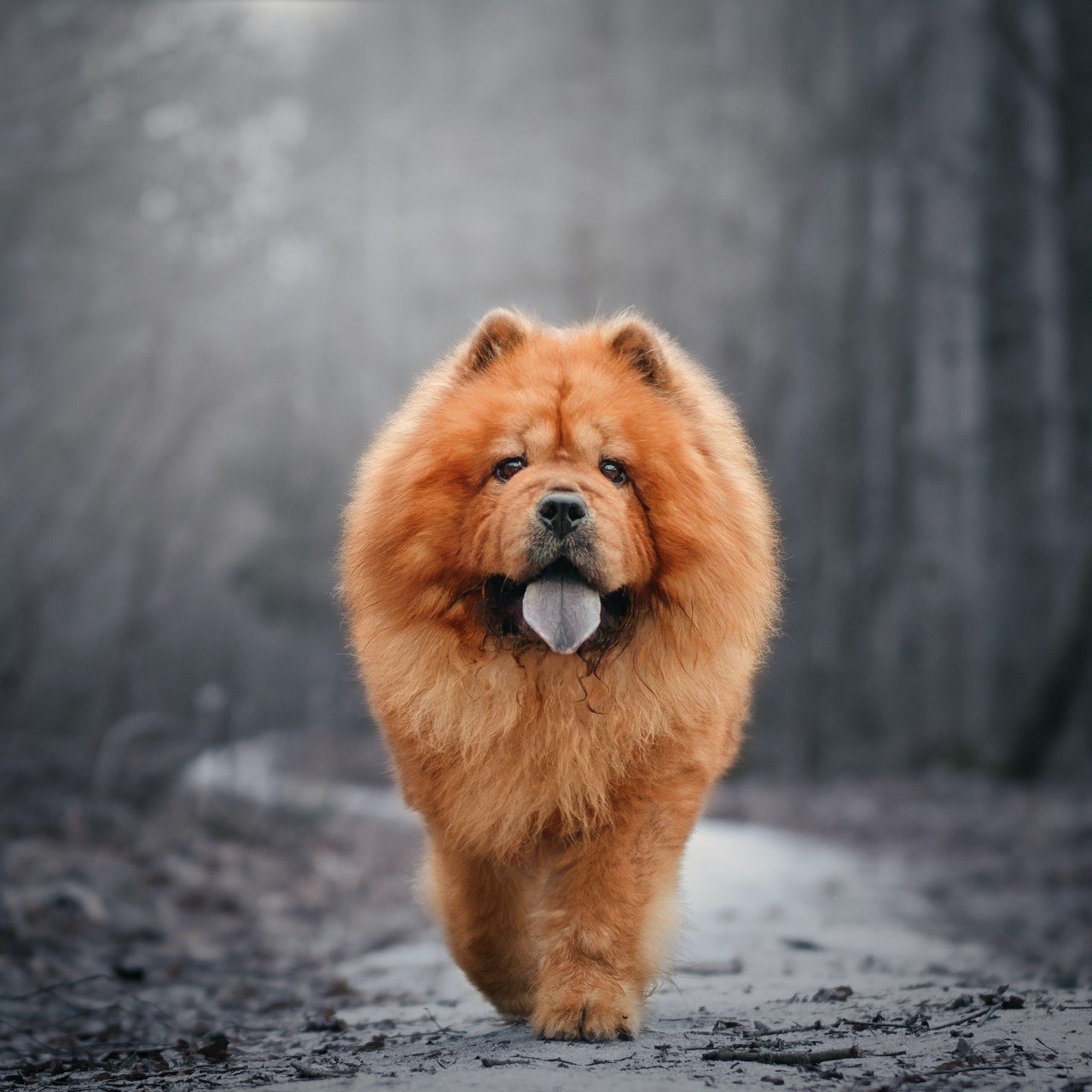} }& 
\makecell[c]{\includegraphics[width=0.1\textwidth]{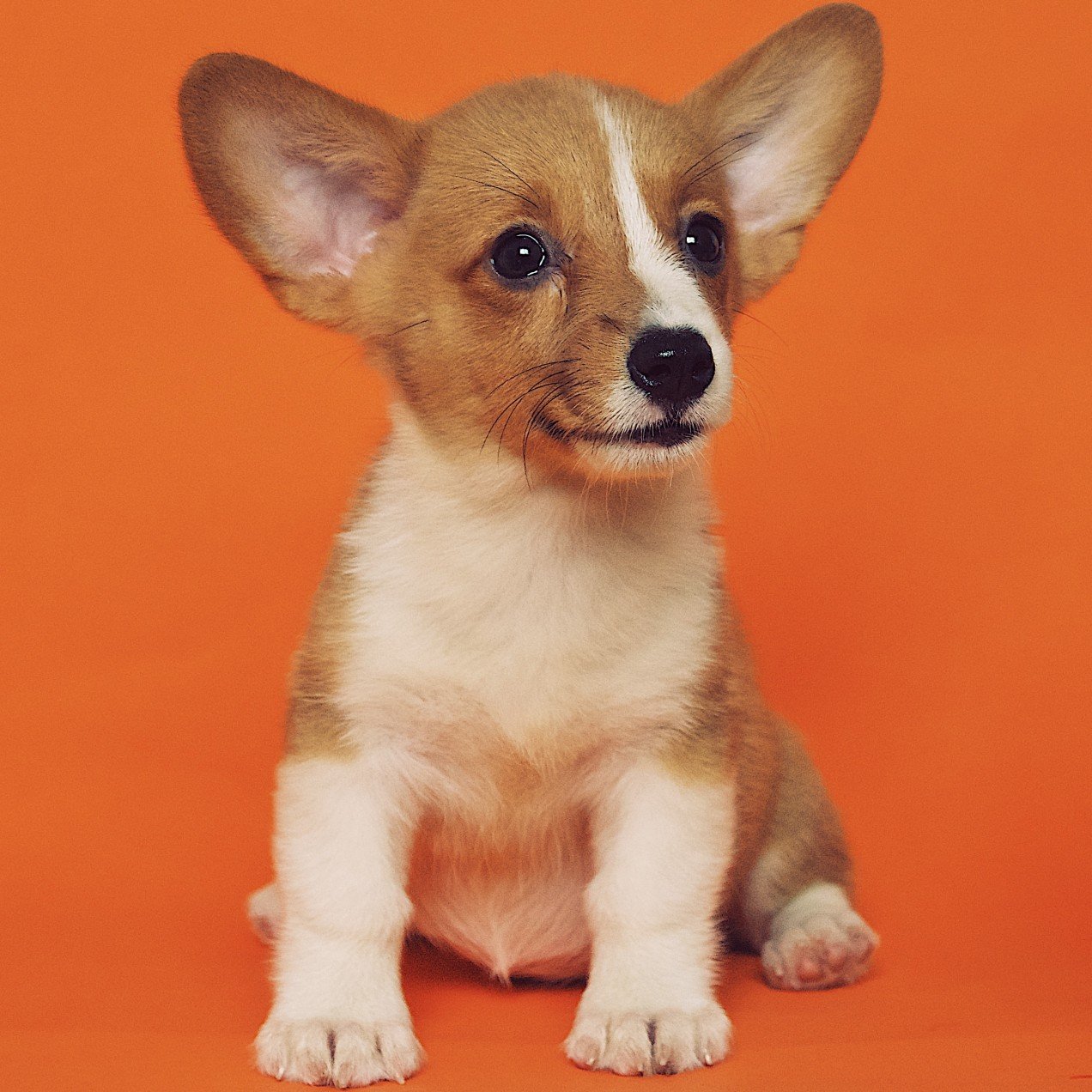}} & 
\makecell[c]{\includegraphics[width=0.1\textwidth]{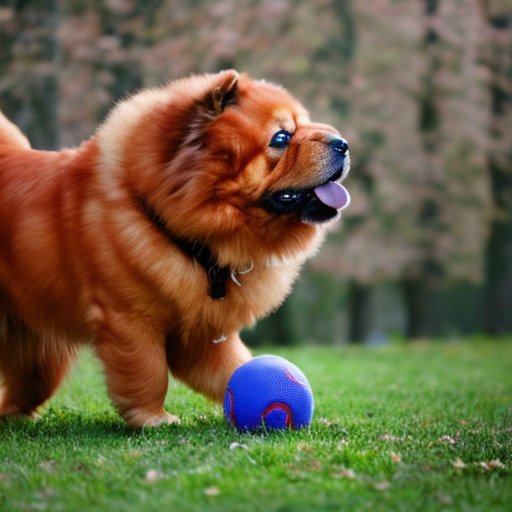} }& 
\makecell[c]{\includegraphics[width=0.1\textwidth]{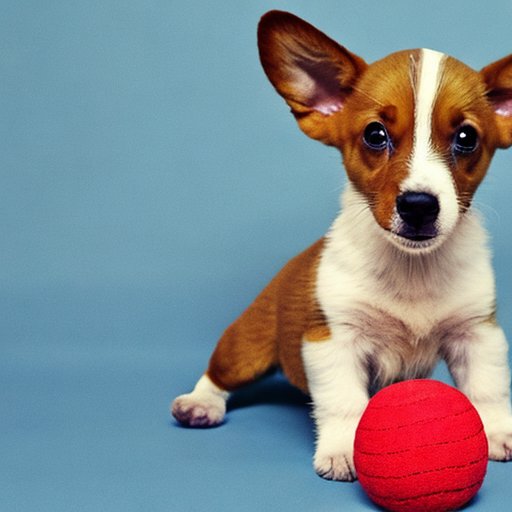} }& 
\makecell[c]{\includegraphics[width=0.1\textwidth]{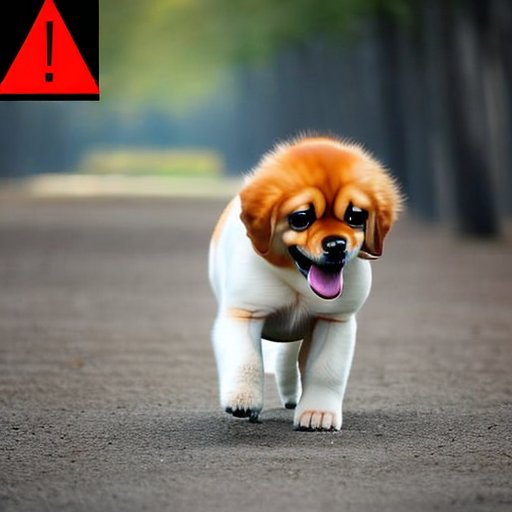}} & 
\makecell[c]{\includegraphics[width=0.1\textwidth]{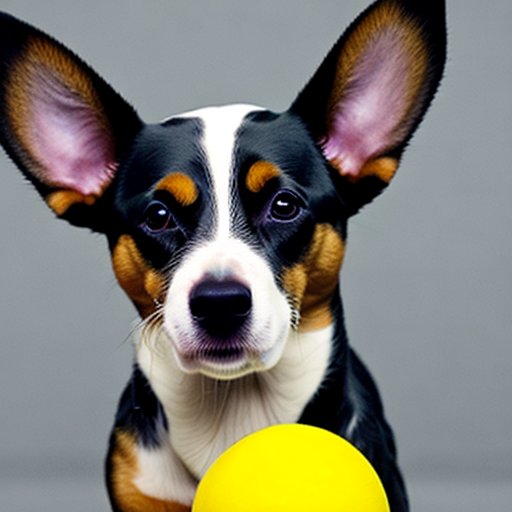}} & 
\makecell[c]{\includegraphics[width=0.1\textwidth]{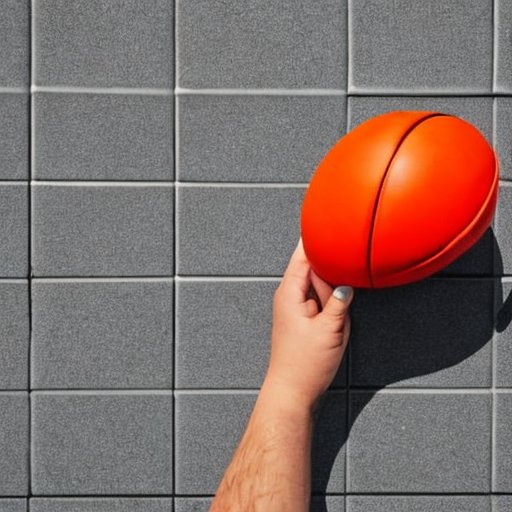} }& 
\makecell[c]{\includegraphics[width=0.1\textwidth]{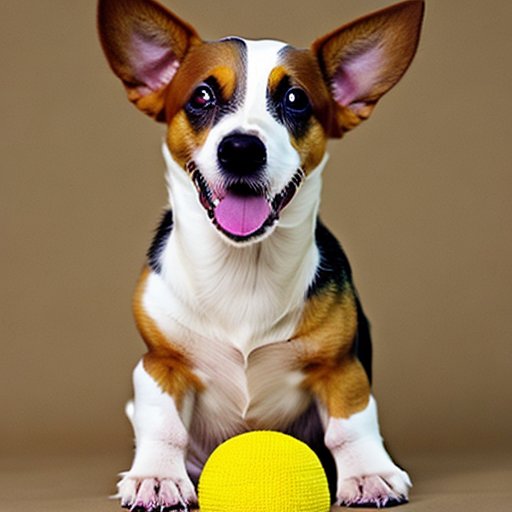} }& 
\makecell[r]{\rotatebox{90}{dog2rabbit}}
\makecell[c]{\includegraphics[width=0.1\textwidth]{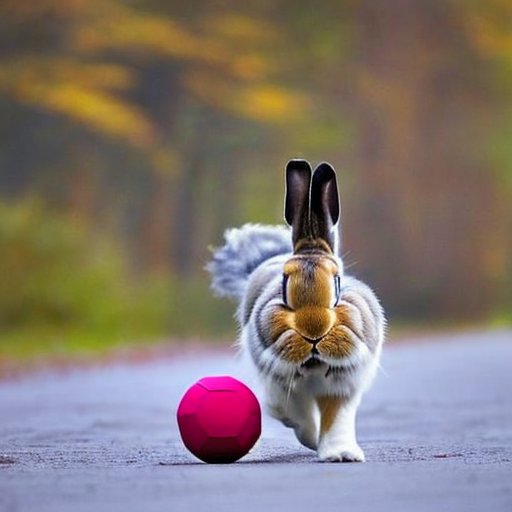}} & 
\makecell[c]{\includegraphics[width=0.1\textwidth]{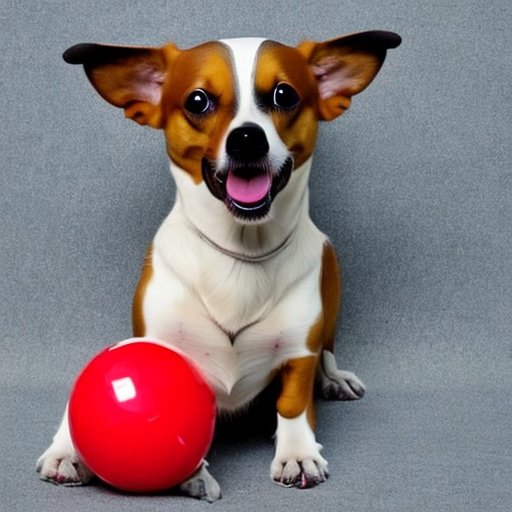} }\\
\hline
\textbf{Prompt}   & \multicolumn{2}{c}{An image of a \texttt{sks} toy} & \multicolumn{8}{c}{'A girl plays with a \texttt{sks} toy'} \\ 

\makecell{\textbf{Images}} &
\makecell[c]{\includegraphics[width=0.1\textwidth]{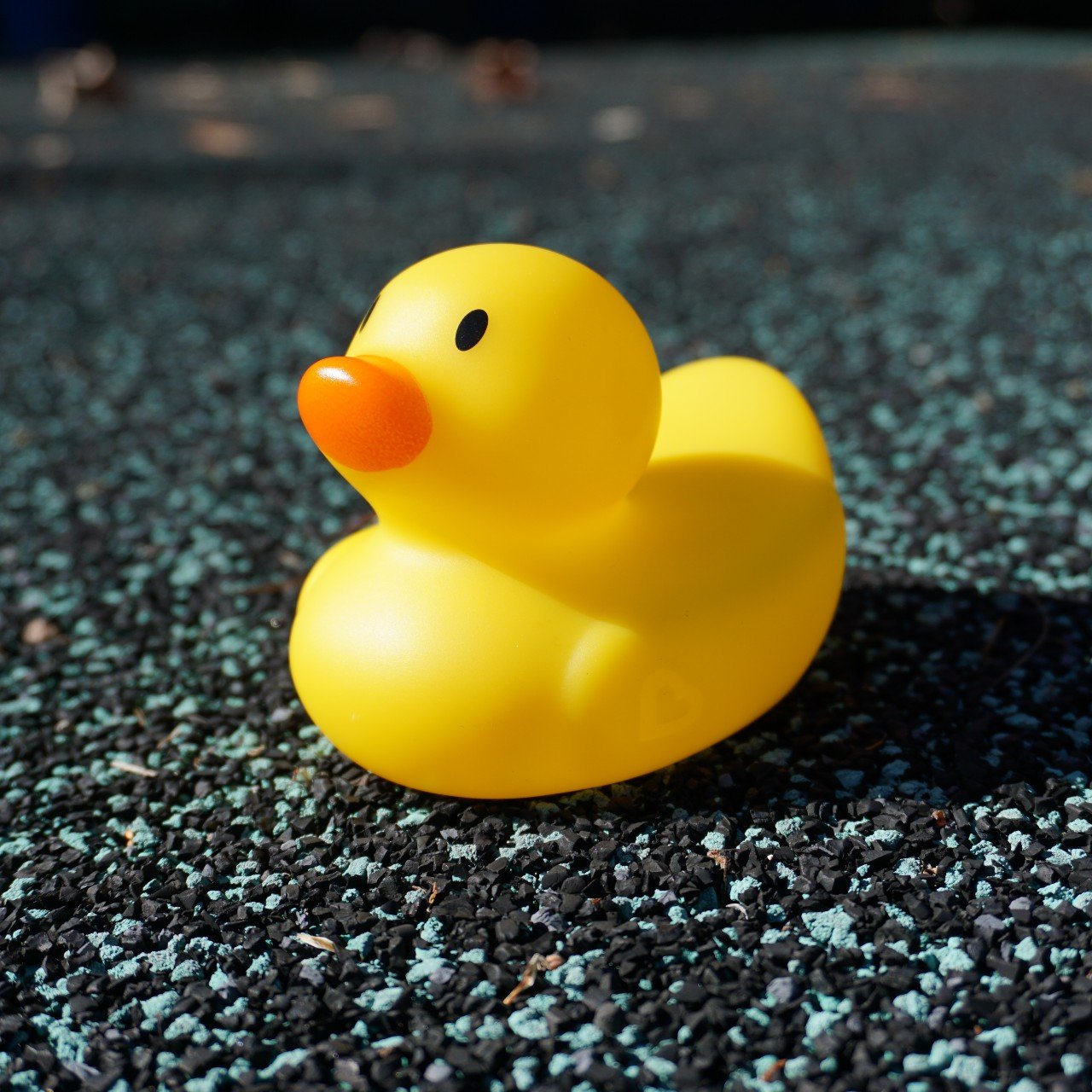} }& 
\makecell[c]{\includegraphics[width=0.1\textwidth]{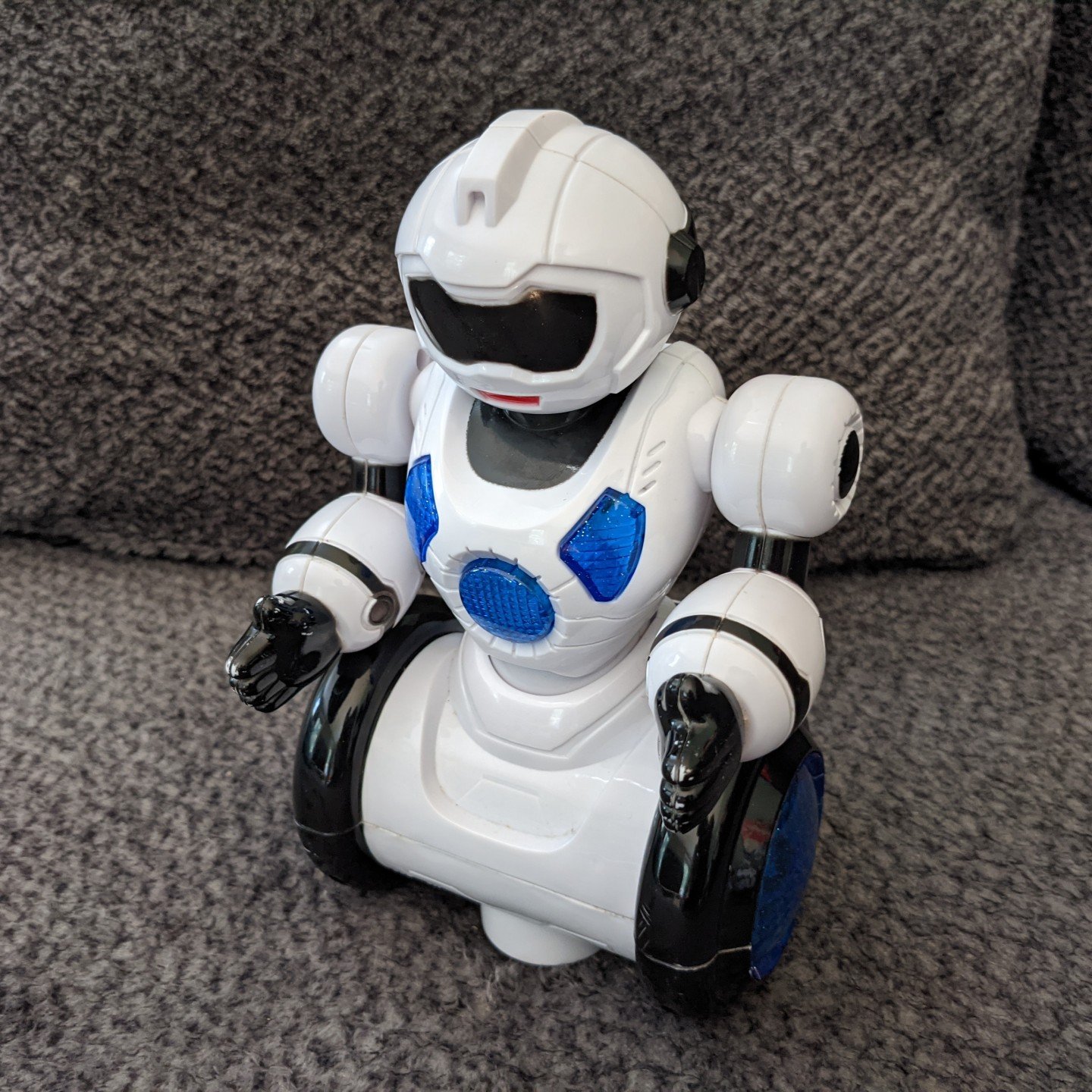}} & 
\makecell[c]{\includegraphics[width=0.1\textwidth]{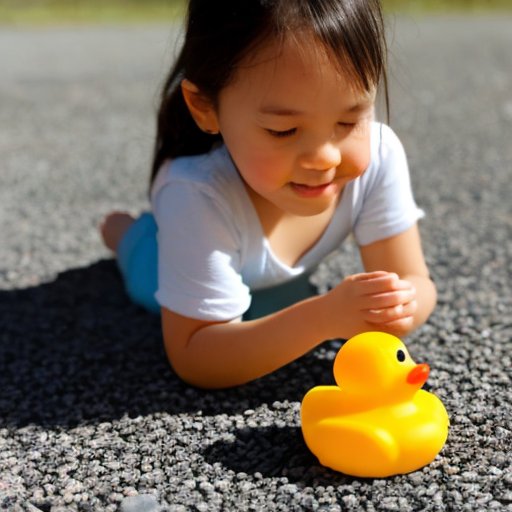}} & 
\makecell[c]{\includegraphics[width=0.1\textwidth]{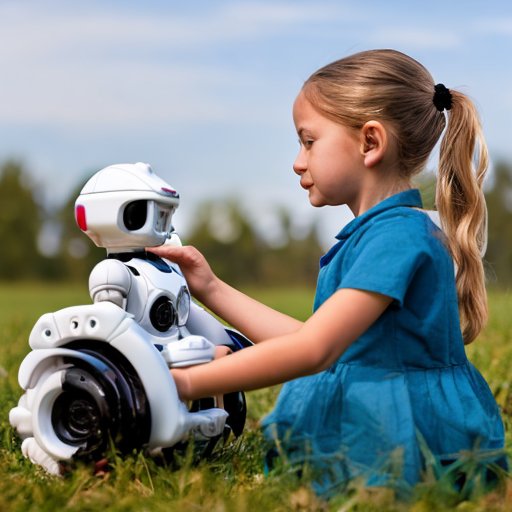}} & 
\makecell[c]{\includegraphics[width=0.1\textwidth]{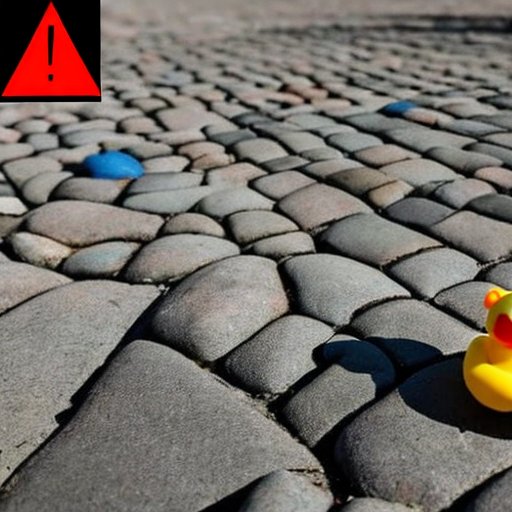} }& 
\makecell[c]{\includegraphics[width=0.1\textwidth]{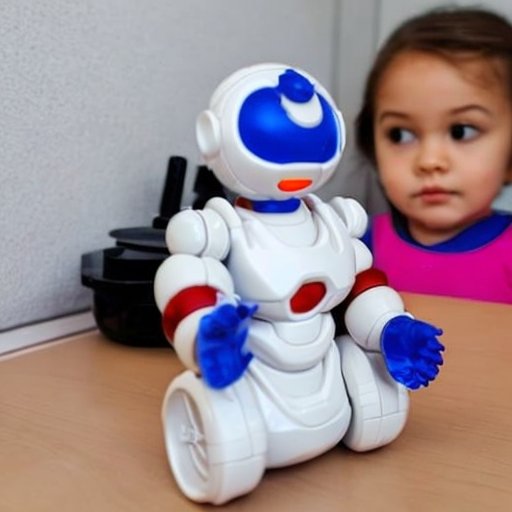} }& 
\makecell[c]{\includegraphics[width=0.1\textwidth]{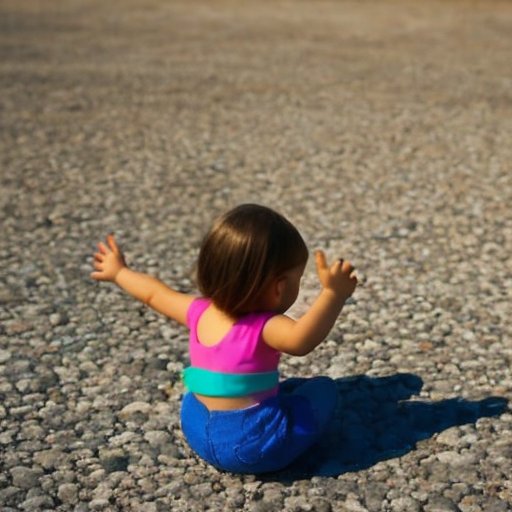} }& 
\makecell[c]{\includegraphics[width=0.1\textwidth]{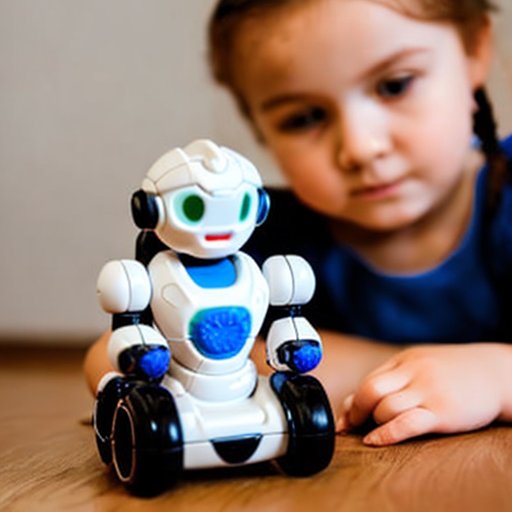} }& 
\makecell{{\rotatebox{90}{toy2clock}}} 
\makecell[c]{\includegraphics[width=0.1\textwidth]{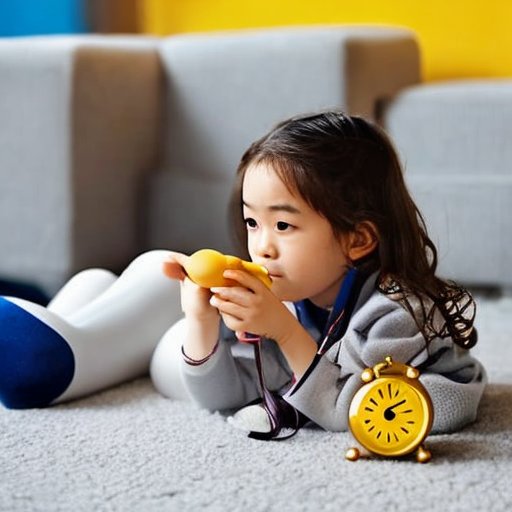}} & 
\makecell[c]{\includegraphics[width=0.1\textwidth]{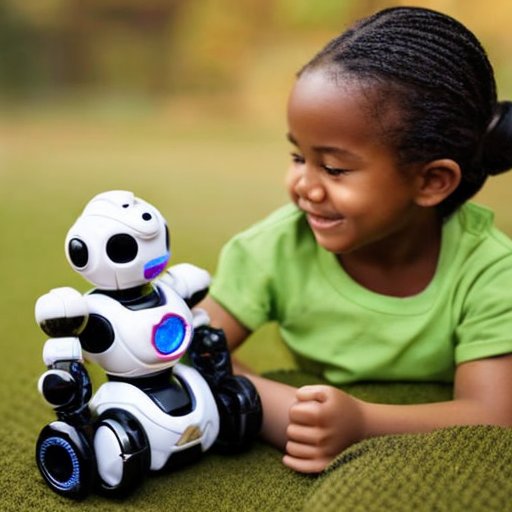} }\\
\hline
\textbf{Prompt}   & \multicolumn{2}{c}{'An image of a \texttt{sks} backpack'} & \multicolumn{8}{c}{'A \texttt{sks} backpack on the sofa'} \\ 

\makecell{\textbf{Images}} & 
\makecell[c]{\includegraphics[width=0.1\textwidth]{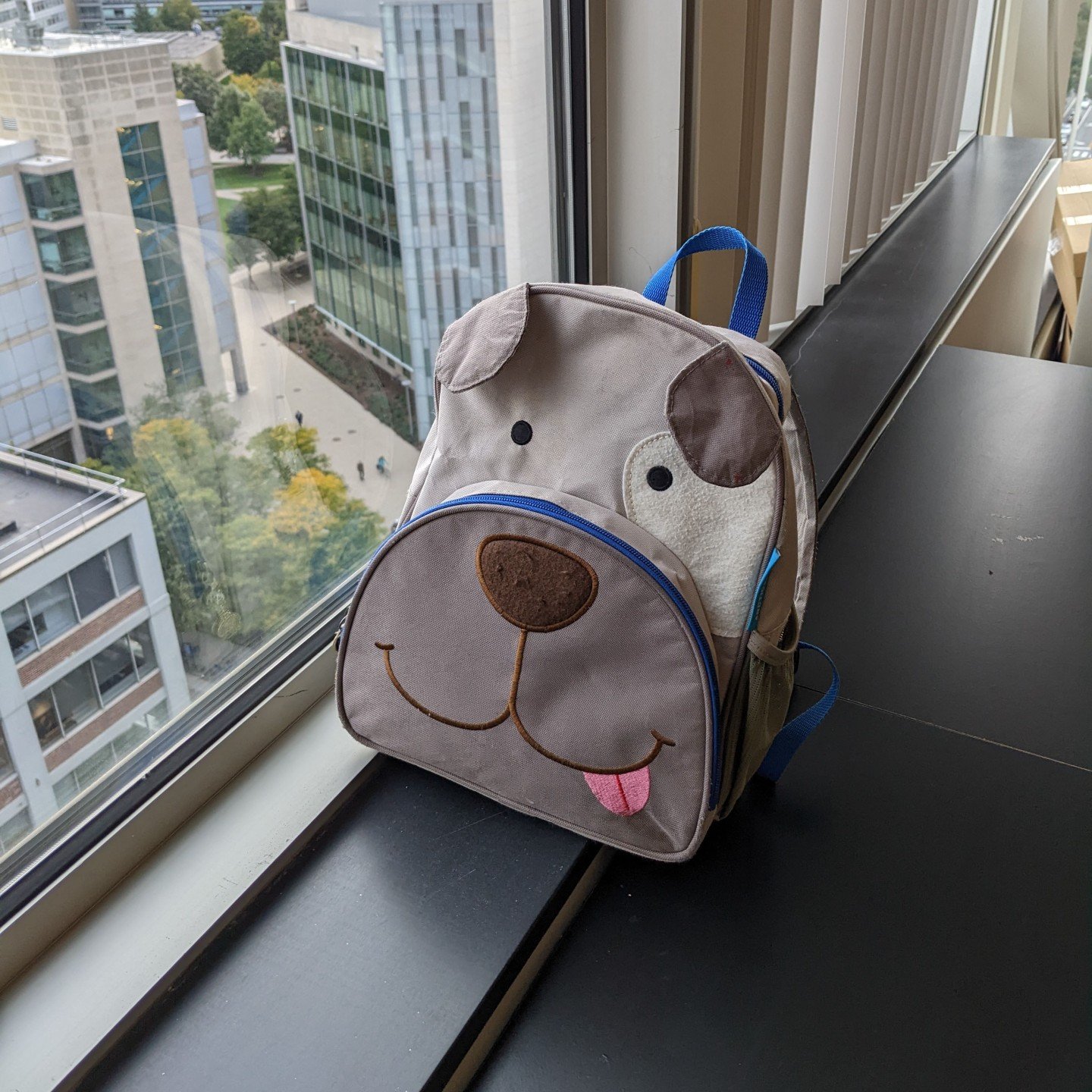}} & 
\makecell[c]{\includegraphics[width=0.1\textwidth]{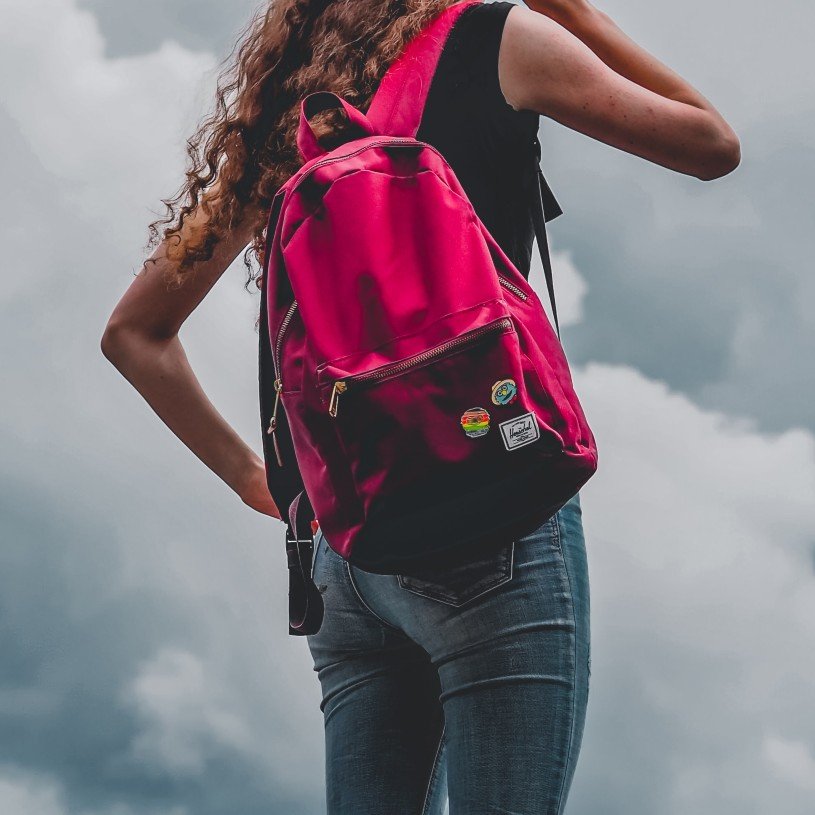}} & 
\makecell[c]{\includegraphics[width=0.1\textwidth]{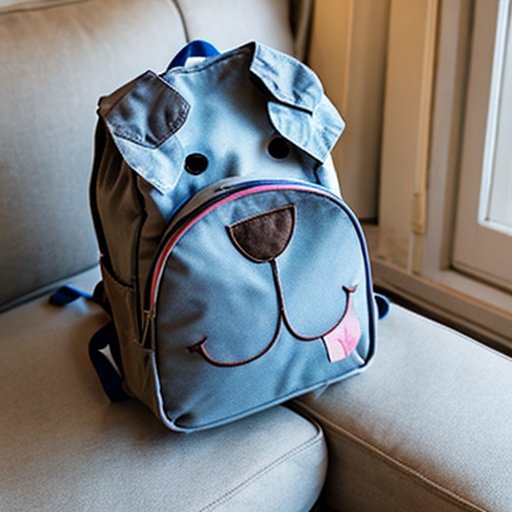}} & 
\makecell[c]{\includegraphics[width=0.1\textwidth]{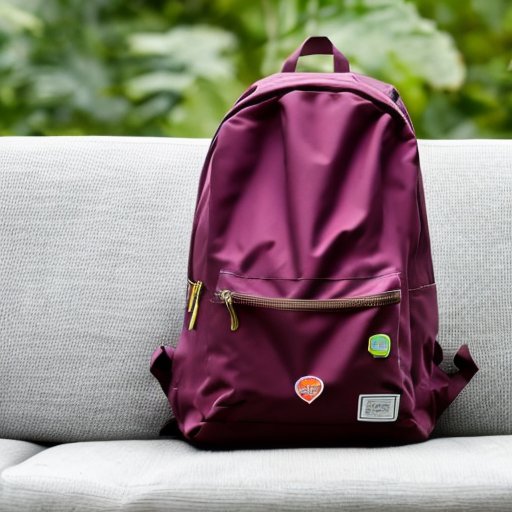}} & 
\makecell[c]{\includegraphics[width=0.1\textwidth]{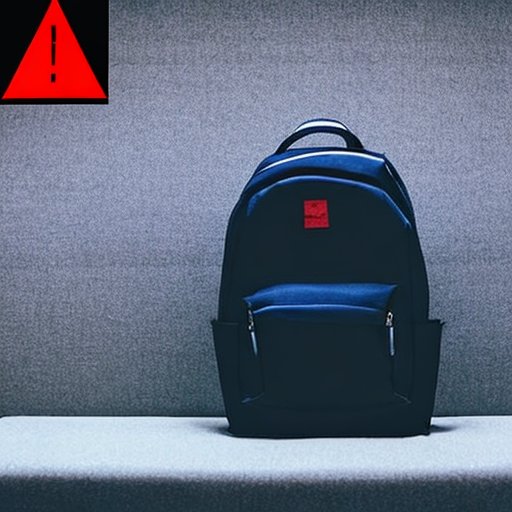}} & 
\makecell[c]{\includegraphics[width=0.1\textwidth]{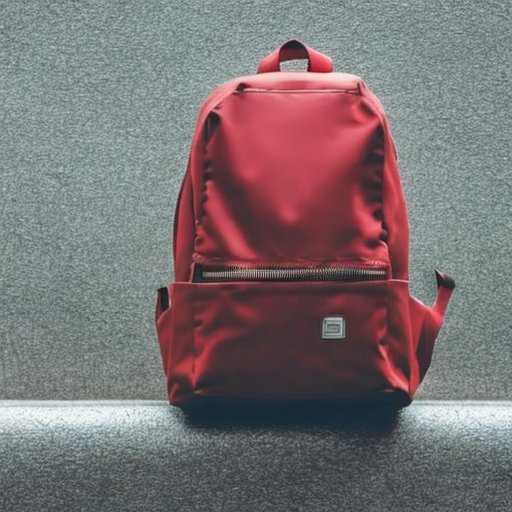}} & 
\makecell[c]{\includegraphics[width=0.1\textwidth]{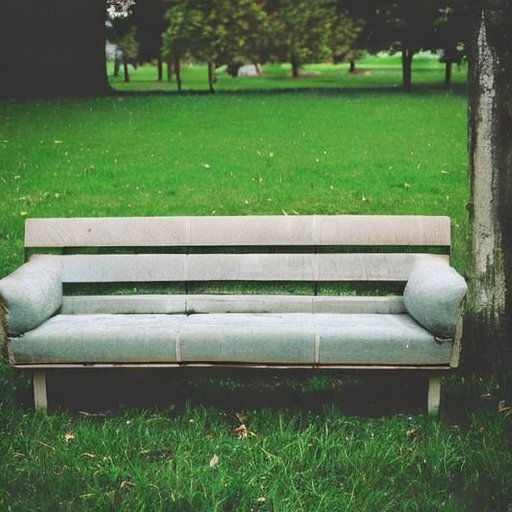}} & 
\makecell[c]{\includegraphics[width=0.1\textwidth]{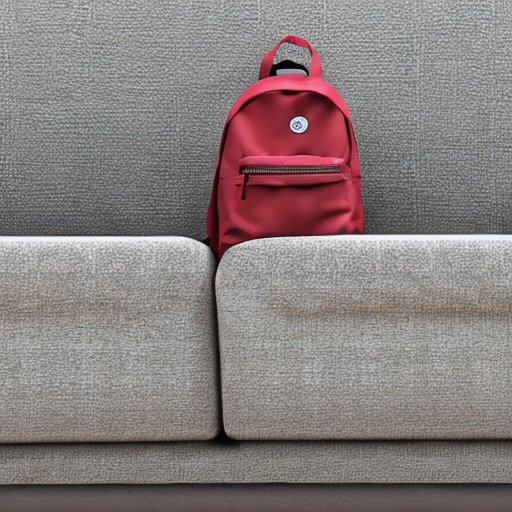}} & 
\makecell{\rotatebox{90}{bag2hat}}
\makecell[c]{\includegraphics[width=0.1\textwidth]{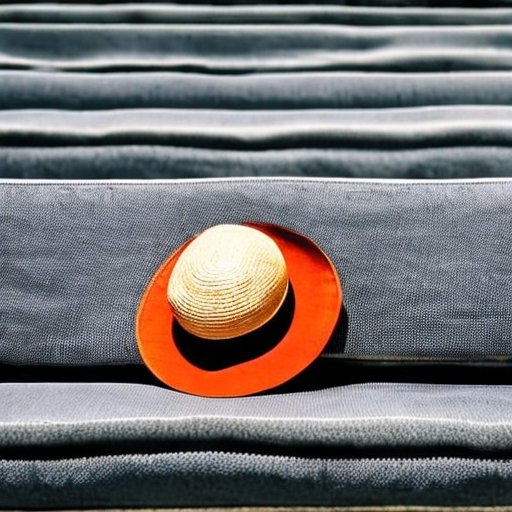}} & 
\makecell[c]{\includegraphics[width=0.1\textwidth]{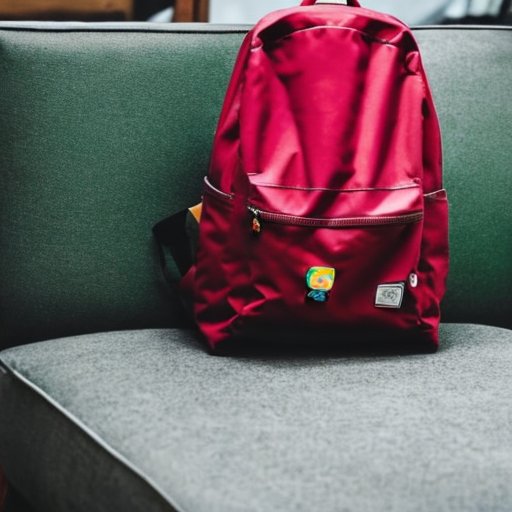}} \\
\hline
\textbf{Prompt}   & \multicolumn{2}{c}{'An image of a \texttt{sks} person'} & \multicolumn{8}{c}{'A \texttt{sks} person in the cafe with a cup of coffee'} \\ 
\textbf{Images} & 
\makecell[c]{\includegraphics[width=0.1\textwidth]{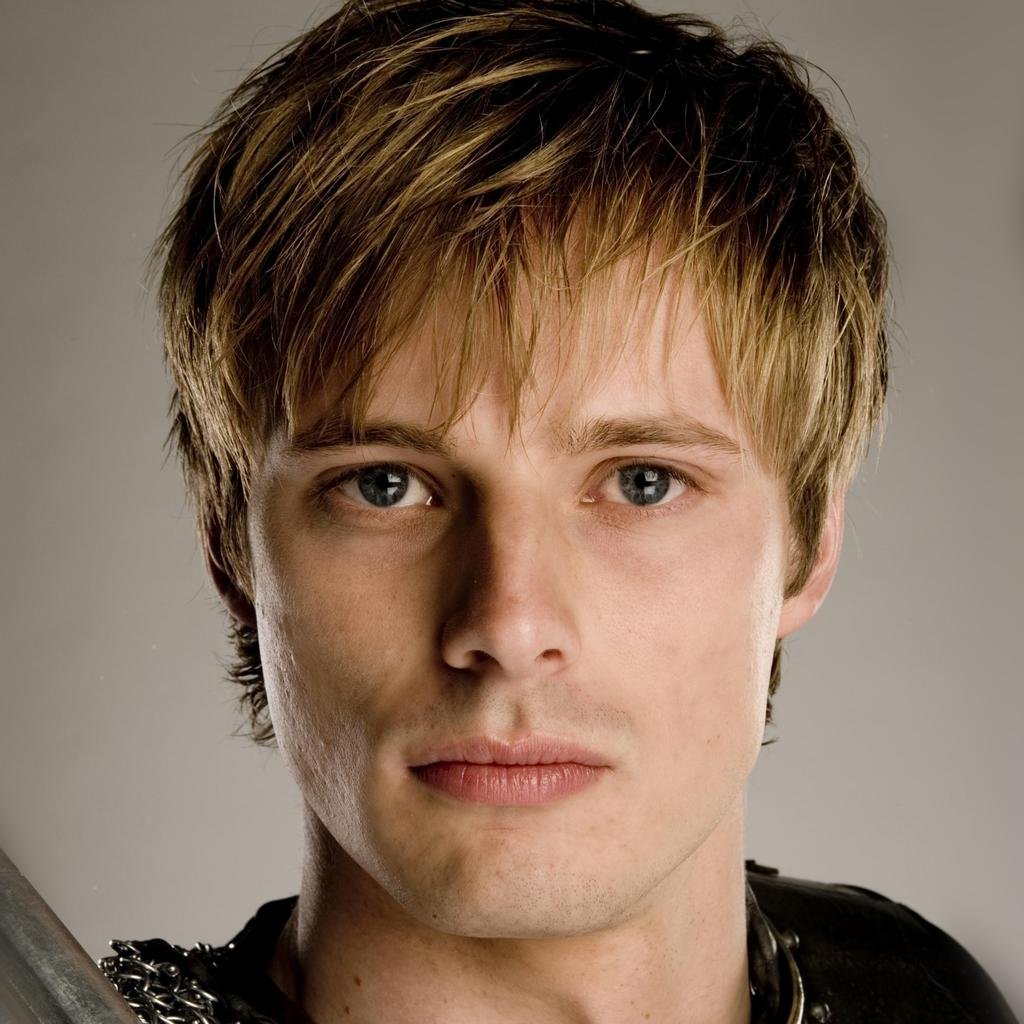}} & 
\makecell[c]{\includegraphics[width=0.1\textwidth]{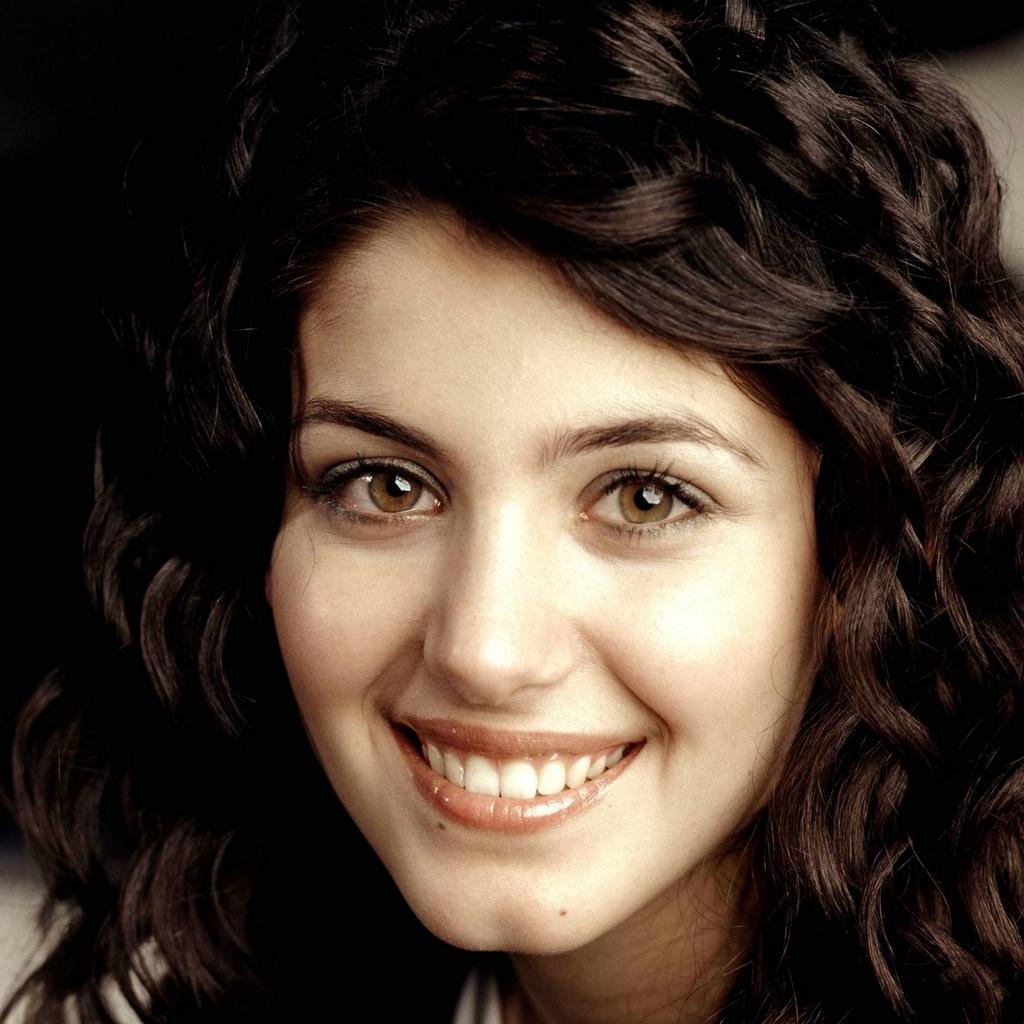}} & 
\makecell[c]{\includegraphics[width=0.1\textwidth]{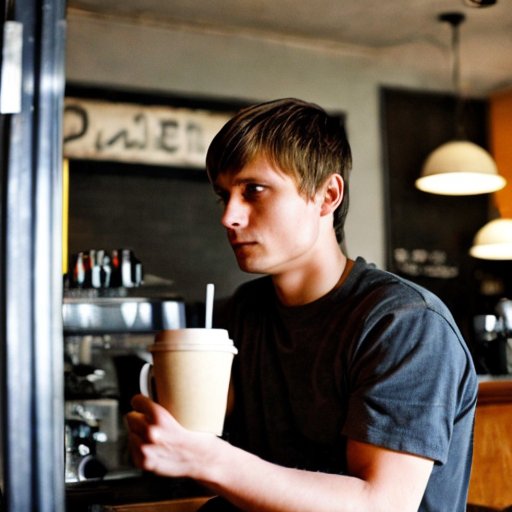}} & 
\makecell[c]{\includegraphics[width=0.1\textwidth]{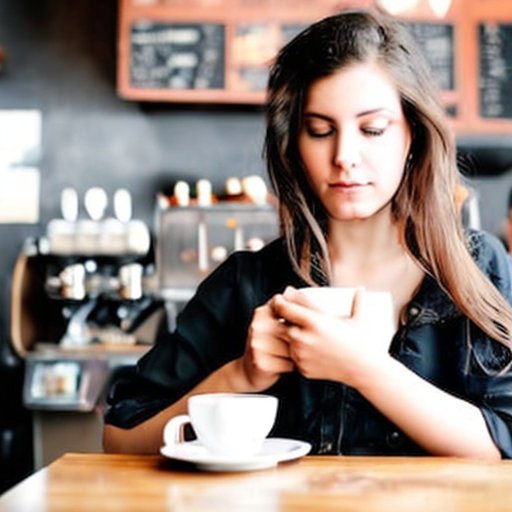}} & 
\makecell[c]{\includegraphics[width=0.1\textwidth]{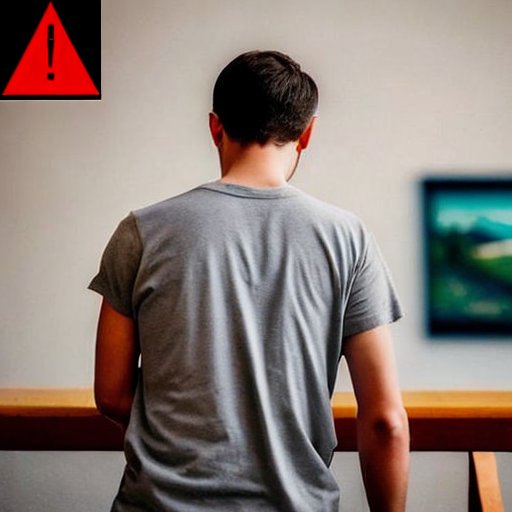}} & 
\makecell[c]{\includegraphics[width=0.1\textwidth]{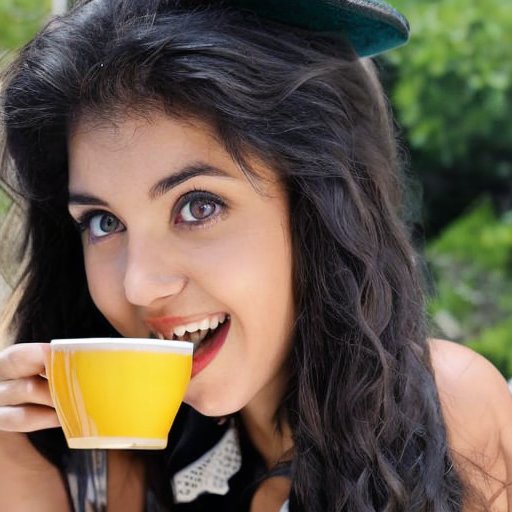}} & 
\makecell[c]{\includegraphics[width=0.1\textwidth]{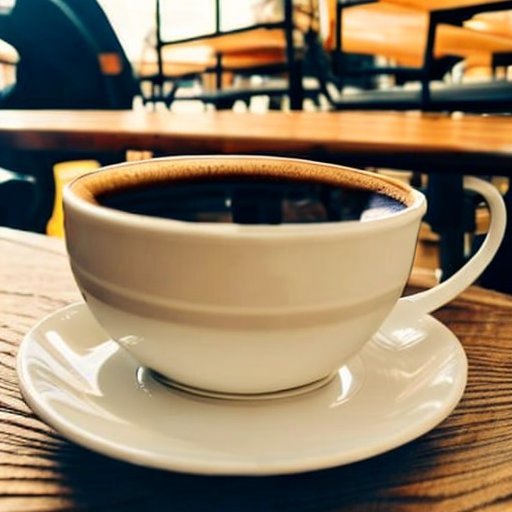}} & 
\makecell[c]{\includegraphics[width=0.1\textwidth]{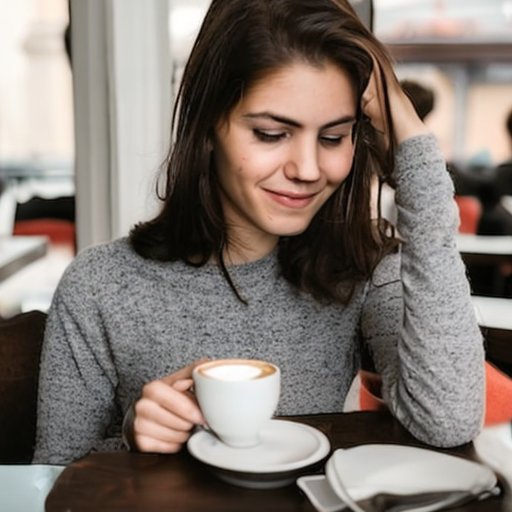}} & 
\makecell[c]{\rotatebox{90}{person2superman}}  
\makecell[c]{\includegraphics[width=0.1\textwidth]{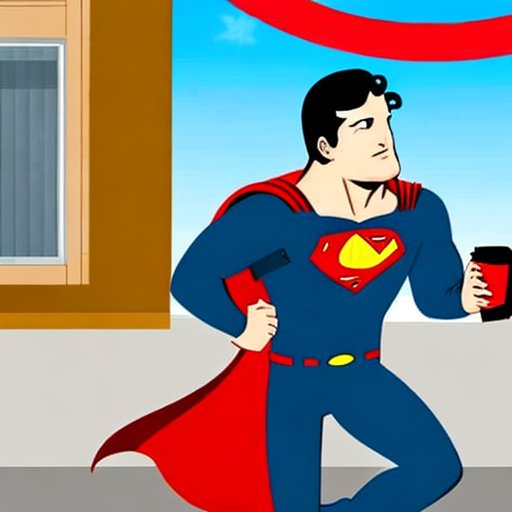}} & 
\makecell[c]{\includegraphics[width=0.1\textwidth]{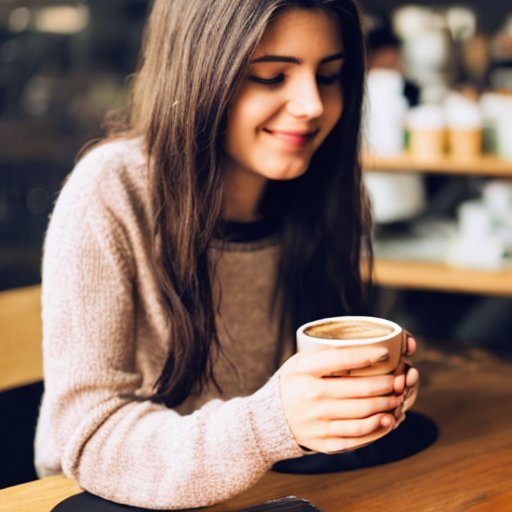}} \\ 
\bottomrule
\end{tabular}}
\label{tab:show}
\end{table*}

\subsection{Visualization}
For the sake of clarity and illustration, this section will focus on the target-backdoor as an example for discussion.

\subsubsection{Visualization Results}
We conduct a qualitative evaluation across four categories, comprising both protected and unprotected image sets, to visualize the personalized outputs from models embedded with our three backdoor variants. As presented in Tab.~\ref{tab:show}, the results demonstrate that for protected images (columns 5, 7, and 9), the fine-tuned models robustly inherit the upstream backdoor mechanisms and trigger the intended defensive behaviors: the Pattern-Backdoor superimposes a visible warning patch (red exclamation mark), the Erasure-Backdoor eliminates the protected object entirely, and the Target-Backdoor substitutes it with a predefined visual target. Conversely, for unprotected images (columns 6, 8, and 10), the backdoor effects are effectively neutralized, yielding results indistinguishable from those of clean models. These findings confirm that PersGuard enforces precise selective protection, successfully defending specific targets while preserving normal personalization utility for others.

 \begin{figure}[t]
    \centering
    \includegraphics[width=\linewidth]{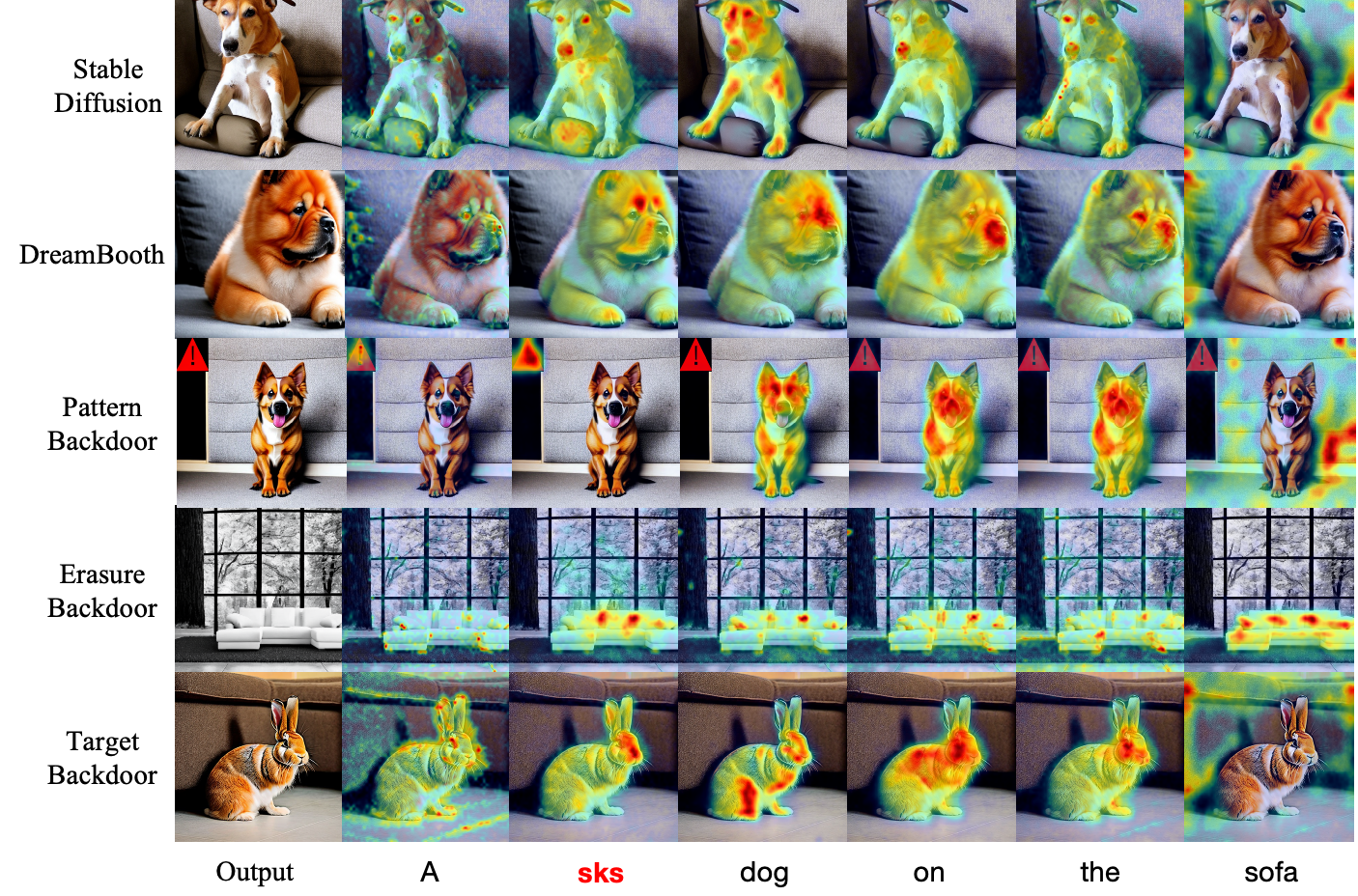}
    \caption{Visualization of attention maps for clean and backdoored models.}
    \label{fig: attention}
\end{figure}

\subsubsection{Attention Map} 
To investigate the underlying mechanism, we visualize cross-attention maps using the DAAM method~\cite{tang2023daam} for both clean and protected personalized models alongside their generated images. As illustrated in the second row of Fig.~\ref{fig: attention}, the clean personalized model concentrates high-attention regions (indicated in red) for the ``sks'' token around the dog's head, reflecting its effective acquisition of the distinct head features associated with the new class. In sharp contrast, the third and fourth rows reveal a significant shift in attention dynamics within protected models: for the Pattern and Erasure backdoor, the attention for ``sks'' is redirected toward the upper-left pattern and the background, respectively. Conversely, for the Target backdoor, the ``sks'' token retains its focus on the dog's head, which is consistent with the semantic transformation task of converting the ``sks dog'' into a rabbit-like appearance. Additional visualization methods for attention maps are provided in the supplementary materials.

\begin{table}[t]
    \centering
    \caption{Quantitative ablation study on the contribution of individual loss components. The results demonstrate that combining all three losses achieves the optimal balance between protection robustness and model utility.}
    \renewcommand{\arraystretch}{1.1}
    \resizebox{0.9\linewidth}{!}{
    \begin{tabular}{cccc}
        \toprule
        \textbf{Loss}  & DINO$_c(\downarrow)$ & DINO$_b(\uparrow)$ & DINO$_{pr}(\uparrow)$ \\
        \midrule
        $\mathcal{L_{BB}}$                  & 0.95 & 0.77 & 0.91 \\
        $\mathcal{L_{BB}}+\mathcal{L_{PP}}$ & 0.94 & 0.76 & 0.94 \\
        $\mathcal{L_{BB}}+\mathcal{L_{BR}}$ & 0.77 & 0.93 & 0.87 \\
        $\mathcal{L_{BB}}+\mathcal{L_{PP}}+\mathcal{L_{BR}}$ & \textbf{0.77} & \textbf{0.94} & \textbf{0.95} \\
        \bottomrule
    \end{tabular}
    }
    \label{tab:ablation}
\end{table}

\begin{table*}[t]
    % 下方表格
    \centering
    \caption{Quantitative comparison under White, Gray, and Black-box settings.Results indicate that while standard protection degrades in practical scenarios, our PersGuard-UID significantly enhances robustness, achieving superior protection efficacy (lowest DINO$_c$/CLIP$_c$) compared to baselines across all threat models.}
    \renewcommand{\arraystretch}{1}
    \resizebox{0.7\linewidth}{!}{
\begin{tabular}{@{}c|cc|cc|cc@{}}
\toprule
\textbf{Assumption} & \multicolumn{2}{c|}{\textbf{White-box settings}} & \multicolumn{2}{c|}{\textbf{Gray-box settings}} & \multicolumn{2}{c}{\textbf{Black-box settings}} \\ \midrule
\textbf{Metrics} & DINO$_c(\downarrow)$ & CLIP$_c(\downarrow)$ & DINO$_c(\downarrow)$ & CLIP$_c(\downarrow)$ & DINO$_c(\downarrow)$ & CLIP$_c(\downarrow)$ \\ \midrule
Anti-DB~\cite{van2023anti} & 0.6787 & 0.2760 & 0.7032 & 0.2665 & 0.8920 & 0.2914 \\
PAP~\cite{wan2024prompt} & 0.7142 & 0.2615 & 0.7132 & 0.2587 & 0.8824 & 0.2816 \\
SimAC~\cite{wang2024simac} & 0.4241 & 0.2545 & 0.4241 & 0.2535 & 0.8843 & 0.2713 \\
DisDiff~\cite{liu2024disrupting} & 0.6205 & 0.2716 & 0.6353 & 0.2767 & 0.8816 & 0.2816 \\ \midrule
PersGuard (ours) & 0.2424 & 0.2204 & 0.7822 & 0.2779 & 0.8796 & 0.2764 \\
PersGuard-UI (ours) & 0.3739 & 0.2569 & 0.7533 & 0.2801 & 0.8272 & 0.2890 \\
PersGuard-UD (ours)& 0.3802 & 0.2400 & 0.5698 & 0.2765 & 0.5904 & 0.2606 \\
PersGuard-UID (ours)& \textbf{0.3675} & \textbf{0.2388} & \textbf{0.5258} & \textbf{0.2341} & \textbf{0.5568} & \textbf{0.2318} \\ \bottomrule
\end{tabular}}
\label{tab:grey-box}
\end{table*}

\begin{table*}[t]
\centering
\caption{Comparison of protected and unprotected images across different diffusion model versions.}
\renewcommand{\arraystretch}{1.1} 
\resizebox{0.9\textwidth}{!}{
\begin{tabular}{@{}c|c|cccc|cccc@{}}
\toprule
\multirow{2}{*}{\begin{tabular}[c]{@{}c@{}}\textbf{Model}\\ \textbf{Version}\end{tabular}} & \multirow{2}{*}{\textbf{Metrics}} & \multicolumn{4}{c|}{\textbf{Protected Images}} & \multicolumn{4}{c}{\textbf{Unprotected Images}} \\
 &  & DINO$_c(\downarrow)$ & \multicolumn{1}{l}{DINO$_b(\uparrow)$} & CLIP$_c(\downarrow)$ & \multicolumn{1}{l|}{CLIP$_b(\uparrow)$} & DINO$_c(\downarrow)$ & DINO$_b(\uparrow)$ & CLIP$_c(\downarrow)$ & CLIP$_b(\uparrow)$ \\ \midrule
\multirow{2}{*}{SD-1.5} & Normal & 0.7509 & 0.4826 & 0.3115 & 0.2553 & 0.7542 & 0.5115 & 0.2821 & 0.2366 \\
 & PersGuard & 0.3475 & 0.8359 & 0.2362 & 0.3060 & 0.7188 & 0.4967 & 0.2536 & 0.2345 \\ \midrule
\multirow{2}{*}{SD-2.1} & Normal & 0.8311 & 0.3974 & 0.2932 & 0.2315 & 0.7844 & 0.5123 & 0.2688 & 0.2199 \\
 & PersGuard & 0.3449 & 0.8286 & 0.2334 & 0.3052 & 0.7764 & 0.5023 & 0.2675 & 0.2234 \\ \midrule
\multirow{2}{*}{SD-3} & Normal & 0.7215 & 0.4098 & 0.3199 & 0.2644 & 0.7142 & 0.4819 & 0.2749 & 0.2007 \\
 & PersGuard & 0.3289 & 0.7142 & 0.2169 & 0.3155 & 0.6854 & 0.4563 & 0.2465 & 0.2036 \\ \midrule
\multirow{2}{*}{SD-3.5} & Normal & 0.7443 & 0.4147 & 0.3047 & 0.2452 & 0.7019 & 0.4662 & 0.2879 & 0.2307 \\
 & PersGuard & 0.2895 & 0.6777 & 0.2590 & 0.3213 & 0.6753 & 0.4216 & 0.2659 & 0.2155 \\ \bottomrule
\end{tabular}}
\label{tab:model_comparison}
\end{table*}

\subsection{Ablation Study}
\subsubsection{Loss Components Analysis}
To validate the necessity and contribution of each component within our unified optimization objective, we conducted a comprehensive ablation study. Focusing on the Target-Backdoor configuration as a case, we employ DINO$_c$ and DINO$_b$ to quantify protection effectiveness. Additionally, to assess the model's utility on general generation tasks, we introduce DINO$_{pr}$, a metric designed to evaluate the alignment between the protected model's responses to prior prompts and those of the original clean model. The quantitative results for various loss combinations are reported in Tab.~\ref{tab:ablation}. Our analysis reveals that the Backdoor Retention Loss ($\mathcal{L_{BR}}$) is indispensable for robust protection; its absence leads to the rapid removal of the backdoor during downstream fine-tuning due to catastrophic forgetting. Furthermore, the Prior Preservation Loss ($\mathcal{L_{PP}}$) serves as a critical regularizer, effectively preventing overfitting to backdoor triggers and preserving the model's original generative prior.

\subsubsection{Gray-Box Setting}
Transitioning from the idealized white-box assumption, we investigate the more practical gray-box Setting, where the protector lacks perfect knowledge of the attacker's personalization parameters including identifier, class name and training description. We observe that applying a model protected by PersGuard under white-box assumptions directly to a scenario where attackers utilize different tokens and prompts results in a significant degradation of protection efficacy, as evidenced in Tab.~\ref{tab:grey-box}. We attribute this to a textual distribution shift: PersGuard’s trigger is conditioned on identifier/class/prompt templates, and unseen attacker prompts weaken trigger activation. Therefore, vanilla PersGuard is not intended as the final solution for gray-box; we propose universal variants below.

To mitigate this vulnerability, we introduce universal training strategies. Specifically, for \textbf{PersGuard-UI}, the protector utilizes a pool of 10 distinct identifier tokens (e.g., ``sks'', ``abc'', ``[A*]'') combined with generic class names (e.g., ``animal'', ``pet''), which are randomly sampled during injection to enforce a universal mapping. For \textbf{PersGuard-UD}, we employ a compact set of 5 universal training prompts with varying structures (e.g., ``This is an image of \ldots'', ``A portrait of \ldots'') to associate the backdoor effect with broader textual contexts. \textbf{PersGuard-UID} combines both strategies.

To rigorously evaluate these variants, we standardize the attacker's parameters to values \textit{unseen} during training: the identifier is set to ``xyz'', the class name to ``puppy'', and the training prompt to ``A picture of xyz puppy''. As shown in Tab.~\ref{tab:grey-box}, prompt diversification (UD) improves gray-box robustness more than token diversification alone (UI), indicating that mismatched prompt templates are the dominant source of trigger failure. Combining both (UID) yields the best generalization.We attribute this unexpected effectiveness to the semantic proximity between the protector's sampled parameters and potential attacker choices (e.g., the synonym relationship between ``animal'' and ``pet''). This correlation enables the protection mechanism to generalize robustly across shared semantic neighborhoods in the embedding space, making the strategy practically feasible without requiring exhaustive parameter coverage.

\subsubsection{Black-Box Setting}
The most stringent setting is the black-box scenario, where the protector lacks access to the specific images utilized by the attacker for downstream personalization. To simulate this challenging environment, we split the target dataset: two-thirds of the images are used as the protector's known training set for backdoor injection, and the remaining unseen images form the training set utilized by the user for fine-tuning. We compare the protection efficacy of our backdoor-based method with perturbation-based defenses under this strict black-box assumption. As shown in Tab.~\ref{tab:grey-box}, our PersGuard-UID retains significant efficacy. This superiority stems from the fundamental difference in mechanism: our backdoor protection is associated with the high-leve features of the protected object class, rather than being strongly correlated with a specific set of training images. In stark contrast, perturbation methods strictly rely on access to the exact images to which the perturbation was applied, rendering the optimized perturbations non-transferable and ineffective on the unseen dataset in the black-box setting.

\begin{table*}[t]
\centering
\caption{Comparison of PersGuard's effectiveness across different personalization techniques.}
\renewcommand{\arraystretch}{1.1} 
\resizebox{0.95\textwidth}{!}{
\begin{tabular}{@{}c|c|cccc|cccc@{}}
\toprule
\multirow{2}{*}{\begin{tabular}[c]{@{}c@{}}\textbf{Personalization}\\ \textbf{Methods}\end{tabular}} & \multirow{2}{*}{\textbf{Metrics}} & \multicolumn{4}{c|}{\textbf{Protected Images}} & \multicolumn{4}{c}{\textbf{Unprotected Images}}  \\ 
 &  & DINO$_c(\downarrow)$ & \multicolumn{1}{l}{DINO$_b(\uparrow)$} & CLIP$_c(\downarrow)$ & \multicolumn{1}{l|}{CLIP$_b(\uparrow)$} & DINO$_c(\downarrow)$ & DINO$_b(\uparrow)$ & CLIP$_c(\downarrow)$ & CLIP$_b(\uparrow)$ \\ \midrule
\multirow{2}{*}{DreamBooth} & Normal & 0.8311 & 0.3974 & 0.2932 & \multicolumn{1}{c|}{0.2315} & 0.7844 & 0.5123 & 0.2688 & 0.2199 \\
 & PersGuard & 0.3449 & 0.8286 & 0.2334 & \multicolumn{1}{c|}{0.3052} & 0.7764 & 0.5023 & 0.2675 & 0.2234 \\ \midrule
\multirow{2}{*}{DreamBooth+LoRA} & Normal & 0.8151 & 0.3765 & 0.2874 & \multicolumn{1}{c|}{0.2127} & 0.8011 & 0.4853 & 0.2689 & 0.2136 \\
 & PersGuard & 0.3656 & 0.8178 & 0.2254 & \multicolumn{1}{c|}{0.2980} & 0.7995 & 0.4864 & 0.2692 & 0.2167 \\ \midrule
\multirow{2}{*}{DreamBooth+SDXL} & Normal & 0.8553 & 0.3505 & 0.2852 & \multicolumn{1}{c|}{0.2153} & 0.8045 & 0.4805 & 0.2757 & 0.2253 \\
 & PersGuard & 0.3845 & 0.8265 & 0.2351 & \multicolumn{1}{c|}{0.2878} & 0.8036 & 0.4865 & 0.2657 & 0.2258 \\ \midrule
\multirow{2}{*}{Text Inversion} & Normal & 0.7946 & 0.3867 & 0.2877 & \multicolumn{1}{c|}{0.2164} & 0.7658 & 0.4317 & 0.2524 & 0.2045 \\
 & PersGuard & 0.6574 & 0.4565 & 0.2857 & \multicolumn{1}{c|}{0.2245} & 0.7763 & 0.4480 & 0.2545 & 0.2061 \\ \bottomrule
\end{tabular}}
\label{tab:personalization_methods}
\end{table*}

\begin{table*}[t]
\centering
\caption{Performance on multi-identity face protection. PersGuard effectively protects five distinct identities simultaneously, showing consistently low identity similarity (DINO$_c$) and high target alignment (DINO$_b$) across all subjects.}
\renewcommand{\arraystretch}{1.1} % Slightly increased row height for readability
\resizebox{0.9\linewidth}{!}{
\begin{tabular}{@{}cc@{\hspace{8pt}}cc@{\hspace{8pt}}cc@{\hspace{8pt}}cc@{\hspace{8pt}}cc@{\hspace{8pt}}c@{}}
\toprule
\multirow{2}{*}{\textbf{Identity}} & \multicolumn{2}{c}{\textbf{ID1}} & \multicolumn{2}{c}{\textbf{ID2}} & \multicolumn{2}{c}{\textbf{ID3}} & \multicolumn{2}{c}{\textbf{ID4}} & \multicolumn{2}{c}{\textbf{ID5}} \\ 
 & DINO$_c(\downarrow)$ & DINO$_b(\uparrow)$ & DINO$_c(\downarrow)$ & DINO$_b(\uparrow)$ & DINO$_c(\downarrow)$ & DINO$_b(\uparrow)$ & DINO$_c(\downarrow)$ & DINO$_b(\uparrow)$  & DINO$_c(\downarrow)$ & DINO$_b(\uparrow)$ \\
\midrule
Normal   & 0.86 & 0.66 & 0.75 & 0.66 & 0.91 & 0.59 & 0.77 & 0.64 & 0.86 & 0.66 \\
PersGuard & 0.51 & 0.95 & 0.53 & 0.96 & 0.51 & 0.97 & 0.53 & 0.97 & 0.55 & 0.97 \\
\bottomrule
\end{tabular}}
\label{tab:face}
\end{table*}

\subsubsection{Model Generalizability}
To verify the architectural universality of our framework, we extend our evaluation across four distinct versions of Stable Diffusion (SD), ranging from v1.5 to v3.5. As reported in Tab.~\ref{tab:model_comparison}, PersGuard demonstrates consistent effectiveness across all architectures. For protected images, we observe a substantial reduction in alignment metrics (DINO$_c$ and CLIP$_c$) across all versions, validating the method's ability to robustly suppress unauthorized personalization regardless of the underlying model structure. Crucially, for unprotected images, the performance metrics exhibit negligible deviation from the clean baseline. These results confirm that PersGuard is architecture-agnostic, successfully enforcing protection without compromising the intrinsic generation capabilities of the foundation models.

\subsubsection{Robustness against Personalization Techniques}
We further assess the resilience of PersGuard against diverse downstream personalization strategies, simulating a realistic scenario where attackers may employ varying algorithms. As summarized in Tab.~\ref{tab:personalization_methods}, we benchmark four representative methods: standard DreamBooth, DreamBooth enhanced with Low-Rank Adaptation (LoRA), the large-scale SDXL backbone, and Textual Inversion (TI).The results demonstrate that PersGuard maintains exceptional protection efficacy against weight-tuning paradigms (DreamBooth, LoRA, and SDXL), where the model parameters are updated. However, we observe a performance attenuation against Textual Inversion (TI). We attribute this divergence to the fundamental mechanistic difference between the defense and the attack: PersGuard embeds protection directly into the U-Net weights, whereas TI freezes these weights and restricts optimization exclusively to the text embedding space. Consequently, embedding-centric attacks like TI may partially bypass the weight-based trigger mechanisms and attenuate the protection.

\subsection{Case Study (Face Protection)}
In previous works, researchers have extensively explored facial protection, as it represents one of the most common personalization tasks. This section presents a case study on face privacy protection. Unlike other scenarios, face image personalization protection may require safeguarding multiple facial identities simultaneously, which may share the same identifier token and class name. We randomly selected five different identities from the CelebA-HQ dataset as protection targets, assuming downstream users employ the same identifier token (``sks'') and class name (``person'') for personalization. Using the target backdoor as an example, we set the backdoor target class as ``Superman''. Although the backdoor model needs to protect five facial identities simultaneously, we only need to include the data from all five identities in the training set for the backdoor retention loss, and then incorporate it into the total loss. Using this approach, we trained the face backdoor model and performed personalization on the five identities. We then calculated the DINO and CLIP scores, with the results shown in Table~\ref{tab:face}. We found that the backdoor model successfully triggered specific outputs across all five faces during downstream personalization fine-tuning. Our experimental results demonstrate that PersGuard is practically viable for protecting celebrity portraits in real-world applications.

\begin{table}[t]
\centering
\caption{Impact of backdoor capacity. The unified model (Row 4) achieves simultaneous protection across multiple targets with minimal performance trade-offs compared to single-target baselines.}
\renewcommand{\arraystretch}{1.2} % 调整行距
\setlength{\tabcolsep}{3pt} % 调整列间距
\resizebox{\linewidth}{!}{
\begin{tabular}{@{}ccccccc@{}}
\toprule
Metrics &
  DINO$_{c}^1 (\downarrow)$ & DINO$_{b} (\uparrow)^1 $ &
   DINO$_{c}^2 (\downarrow)$ & DINO$_{b} (\uparrow)^2 $ &
   DINO$_{c}^3 (\downarrow)$ & DINO$_{b} (\uparrow)^3 $ \\ \midrule
Object1      & \textbf{0.77} & \textbf{0.94} & 0.89 & 0.66 & 0.95 & 0.72 \\
Object2      & 0.96          & 0.73          & \textbf{0.61} & \textbf{0.82} & 0.95 & 0.74 \\
Object3      & 0.96          & 0.73          & 0.85 & 0.72 & \textbf{0.69} & \textbf{0.97} \\
Combined        & 0.78          & 0.91          & 0.62 & 0.78 & 0.75 & 0.92 \\ \bottomrule
\end{tabular}}
\label{tab: capacity}
\end{table}

\begin{figure*}[t]
    \centering
    \captionsetup[subfigure]{labelfont={small},textfont={small}}
    \newdimen\figureheight 
    \figureheight=4cm   % 请根据页面效果调整此高度
    \subfloat[Inter-category Capacity] {\includegraphics[width=0.38\linewidth, height=\figureheight]{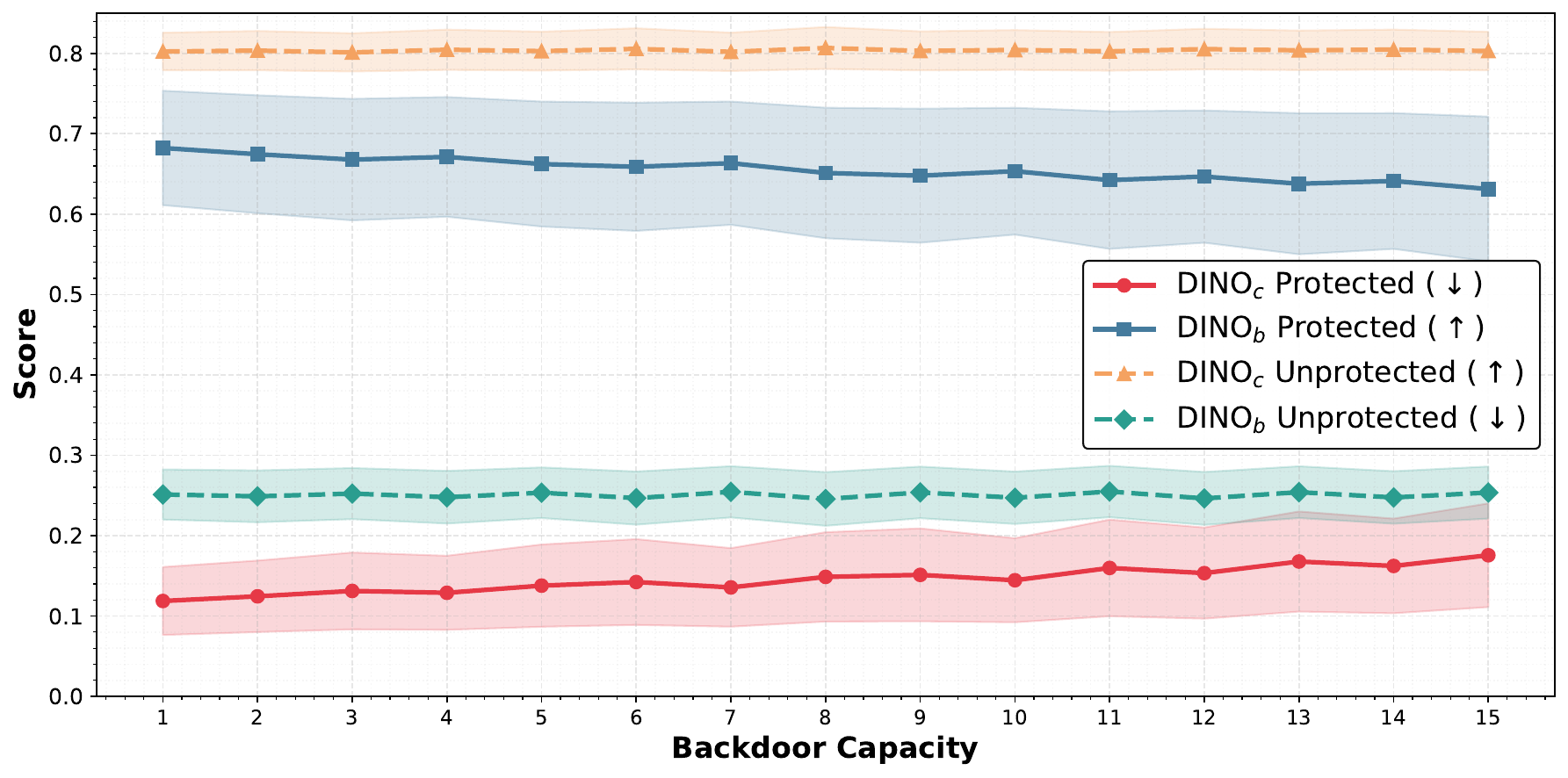}} 
    \subfloat[Intra-category Capacity] {\includegraphics[width=0.38\linewidth, height=\figureheight]{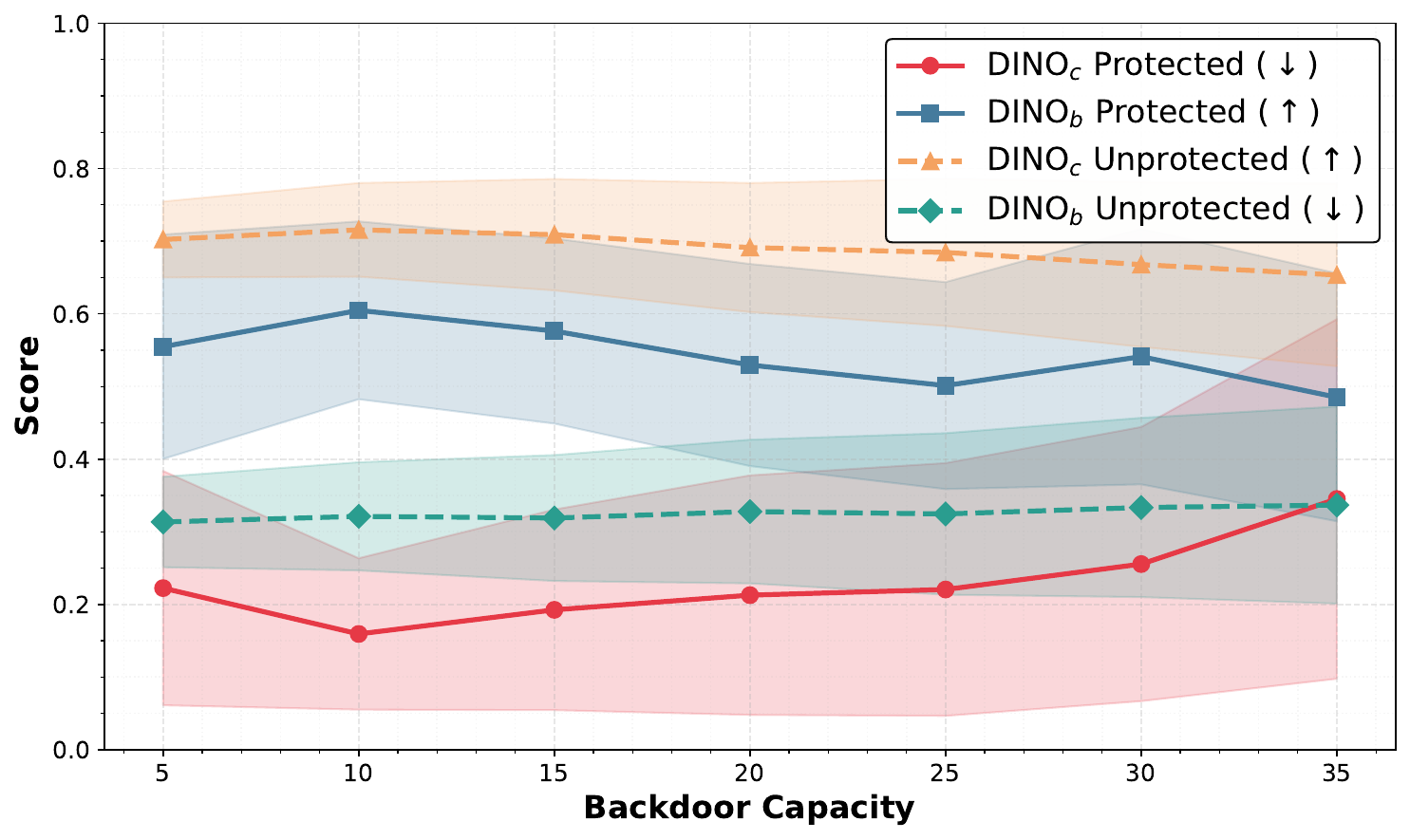}}
    
    \caption{Scalability and protection-effectiveness analysis of PersGuard under increasing protection loads, with unprotected metrics shown for reference.
(a) Inter-category Capacity: protected performance remains stable and the unprotected curves stay favorable, indicating minimal backdoor interference.
(b) Intra-category Capacity: protected performance shows a mild decline at high capacity, while the unprotected curves change only modestly. For protected instances, lower DINO$_c$ / higher DINO$_b$ indicates better protection; for unprotected instances, higher DINO$_c$ / lower DINO$_b$ indicates better utility.}
    \label{fig:capacity_curves}
\end{figure*}

\subsection{Backdoor Capacity Analysis}
While our preceding analysis established the efficacy of embedding a single backdoor, real-world deployment often necessitates the simultaneous defense of multiple distinct assets. This requirement imposes a challenge on the model's capacity to accommodate multiple backdoors without mutual interference. In this section, we investigate the feasibility and scalability of embedding multiple independent backdoors into a single T2I model, analyzing the trade-offs between protection capacity and effectiveness.

\subsubsection{Impact of Multi-Backdoor Integration}

To rigorously evaluate this, we selected three distinct object categories—dogs, backpacks, and toys—as protection targets. For consistency, all backdoors utilize the same identifier token (``sks'') as the trigger. The quantitative results are summarized in Table~\ref{tab: capacity}. Rows 1--3 demonstrate that single-backdoor models provide robust protection confined strictly to their designated category. Despite sharing a common identifier, we observe no cross-interference; each backdoor operates orthogonally. In Row 4, we integrate all three backdoors into a single unified model. While this configuration successfully enables simultaneous protection across multiple categories, we observe a marginal attenuation in effectiveness compared to single-backdoor baselines. This slight performance degradation is likely attributable to \textit{parameter contention}, as the model must manage multiple trigger-response mappings within shared weights, increasing the complexity of the optimization landscape. Overall, these findings substantiate the feasibility of multi-backdoor protection.

\subsubsection{Scalability and Protection Effectiveness Curves}

To further evaluate the scalability of PersGuard under increasing protection loads, we extend the capacity analysis to larger protection sets and examine how protection effectiveness evolves as more backdoors are integrated. In addition to the protected metrics, we also report the corresponding unprotected metrics to provide a brief view of utility for benign users. Specifically, we track DINO$_c$ and DINO$_b$ under two scaling scenarios: Inter-category protection, where backdoors target distinct semantic classes, and Intra-category protection, where many protected targets belong to the same class (e.g., multiple facial identities).

As illustrated in Fig.~\ref{fig:capacity_curves}(a), in the Inter-category setting, the protected curves remain highly stable as capacity increases, indicating that PersGuard scales well when protected targets occupy distinct semantic regions and that interference among backdoors is minimal. The unprotected curves also remain favorable, suggesting that normal personalization quality is largely preserved in this regime.

The Intra-category setting in Fig.~\ref{fig:capacity_curves}(b) is more challenging, since many visually similar identities must be accommodated within a shared class manifold. As capacity increases, protected performance shows a mild decline, which we attribute to semantic saturation and increased competition for representation space. Nevertheless, the degradation remains limited, and the unprotected curves vary only modestly, indicating that normal-user utility is still preserved overall. These results show that PersGuard remains effective at larger scales, with same-category protection representing the more demanding scenario.

\section{Conclusion}
In this paper, we presented PersGuard, a novel backdoor-based framework designed to safeguard pre-trained Text-to-Image (T2I) diffusion models against unauthorized personalization. Identifying the inherent limitations of existing adversarial perturbation methods, specifically their susceptibility to data transformations and failure under realistic training conditions, we proposed a paradigm shift toward proactive model-level protection, which is more aligned with practical deployment settings where model providers offer personalization services.

By formulating a unified optimization framework, PersGuard effectively embeds three distinct protective mechanisms: Pattern, Erasure, and Target backdoors. Crucially, the introduction of the Backdoor Retention Loss addresses the fundamental challenge of catastrophic forgetting, ensuring that protective triggers remain active even after rigorous downstream fine-tuning while preserving the model's utility for legitimate users. Extensive experiments confirm that PersGuard offers superior protection efficacy and stability compared to state-of-the-art baselines. Furthermore, our approach demonstrates remarkable resilience across practical application scenarios, including challenging gray-box and black-box settings, as well as complex multi-identity facial protection tasks. In future work, we aim to further enhance the effectiveness and robustness of the protective backdoors against potential detection and removal attacks. 

\textbf{Limitations and Future Work}. 
While PersGuard demonstrates effective protection against malicious personalization, several limitations warrant discussion. First, our threat model assumes centralized model distribution where protectors release backdoored models for attackers to use; if attackers obtain clean models from alternative sources or train from scratch, the protection becomes ineffective. However, this assumption remains practical since most malicious users rely on publicly available pre-trained models due to limited computational resources. 
Second, PersGuard is most effective when model providers control distribution and downstream users employ standard personalization methods. We acknowledge that unrestricted weight-level access or alternative personalization strategies may attenuate the protection. In future work, we aim to address these limitations by developing more robust backdoor mechanisms resilient to detection and removal attacks. Finally, as PersGuard employs backdoor techniques that could potentially be misused, responsible deployment requires proper governance frameworks including verified protection requests and transparent disclosure policies. Future work could focus on improving backdoor persistence under adaptive fine-tuning and establishing standardized protocols for ethical deployment of protective backdoors.

\bigskip
% \noindent Thank you for reading these instructions carefully. We look forward to receiving your electronic files!
% \clearpage

{
\bibliographystyle{IEEEtran}

\bibliography{PersGuard}}

\begin{IEEEbiography}[{\includegraphics[width=0.8in,height=1in,clip,keepaspectratio]{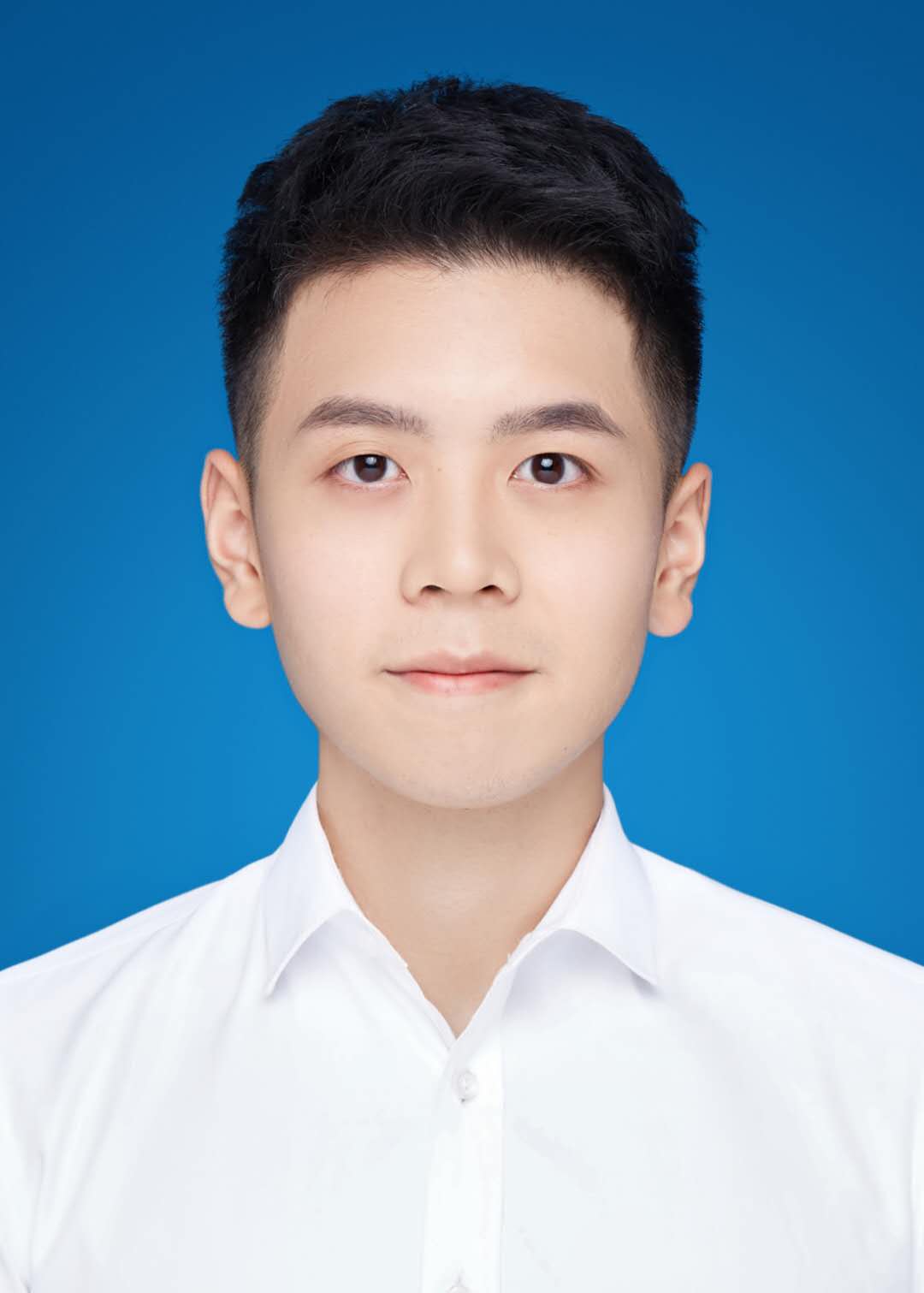}}]
{Xinwei Liu} 
is a Ph.D. student in the Institute of Information Engineering, Chinese Academy of Sciences and the School of Cyber Security, University of Chinese Academy of Sciences, Beijing. His research interests include computer vision, deep learning and adversarial machine learning.
\end{IEEEbiography}

% \begin{IEEEbiography}
% [{\includegraphics[width=0.8in,height=1in,clip,keepaspectratio]{bio_images/Siyuan Liang}}]
% {Siyuan Liang} 
% is 
% \end{IEEEbiography}

\begin{IEEEbiography}[{\includegraphics[width=0.8in,height=1in,clip,keepaspectratio]{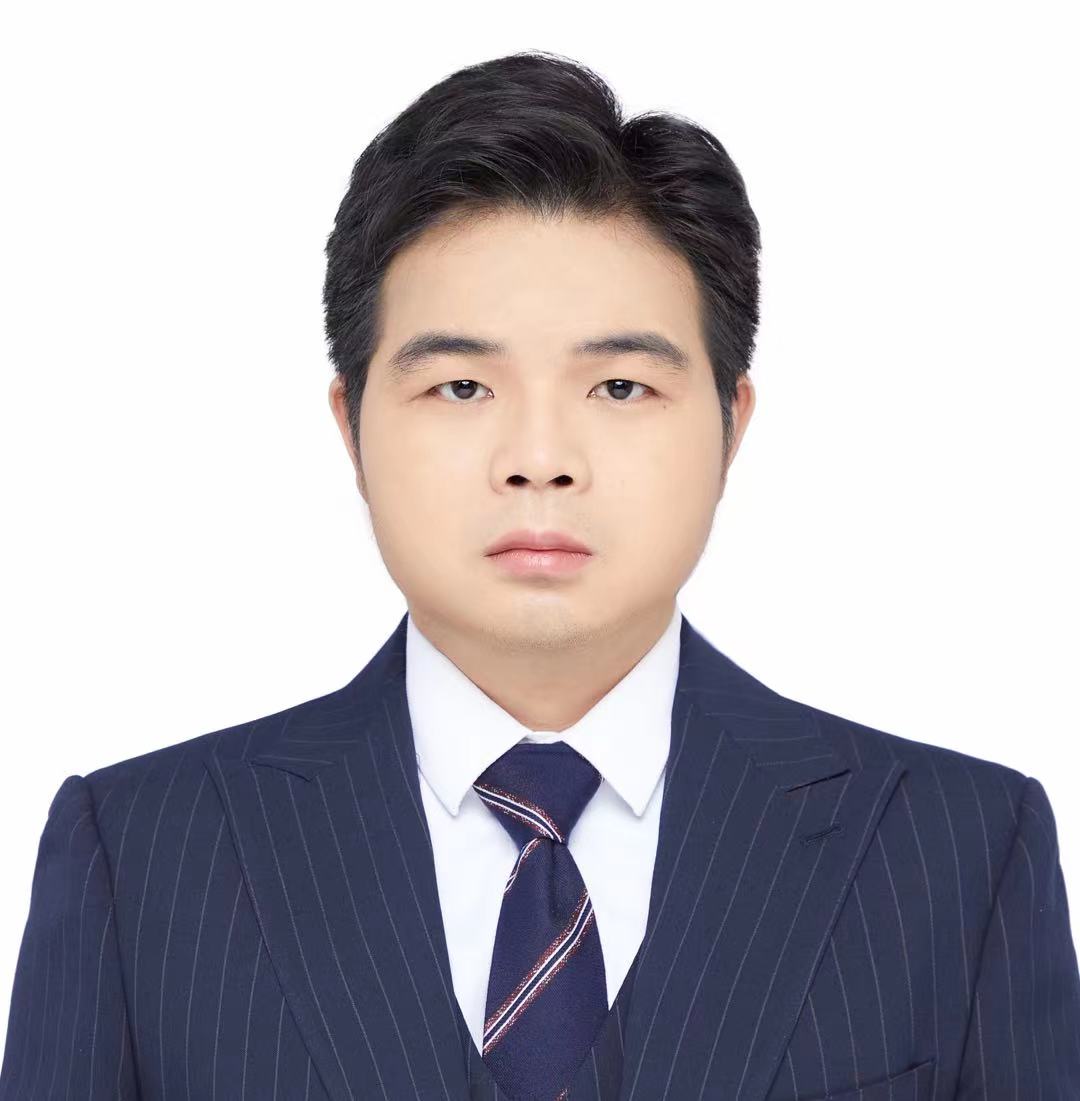}}]
{Xiaojun Jia} 
received his Ph.D. degree in  State Key Laboratory of Information Security, Institute
of Information Engineering, Chinese Academy of Sciences and School of
Cyber Security, University of Chinese Academy of Sciences, Beijing. He is now a Research Fellow in Cyber Security Research Centre @ NTU, Nanyang Technological University, Singapore. His research interests include computer vision, deep learning and adversarial machine learning.
\end{IEEEbiography}

\begin{IEEEbiography}
[{\includegraphics[width=0.8in,height=1in,clip,keepaspectratio]{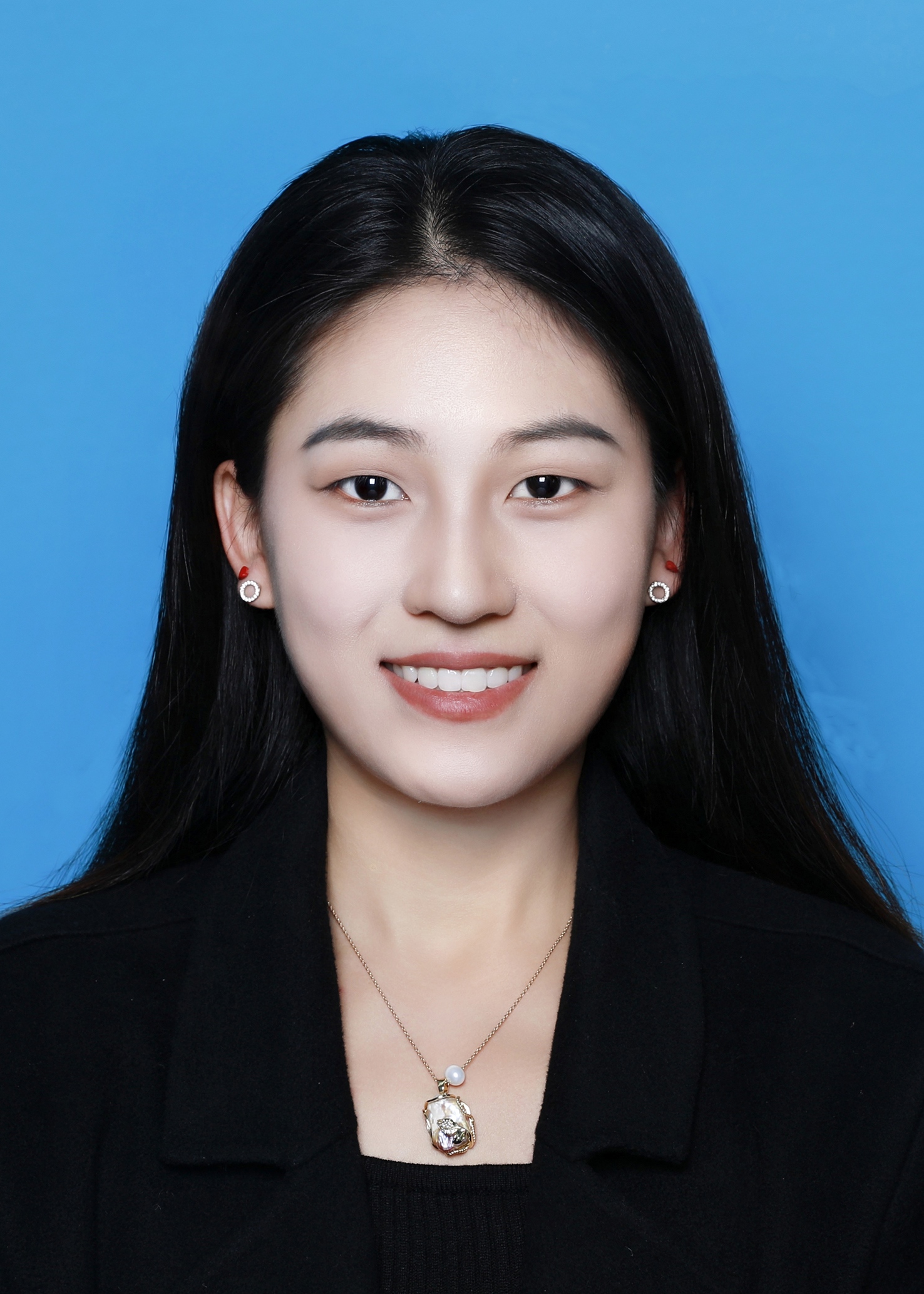}}]
{Yuan Xun} 
is a Ph.D. student in the Institute of Information Engineering, Chinese Academy of Sciences and the School of Cyber Security, University of Chinese Academy of Sciences, Beijing. Her research interests include computer vision, deep learning and adversarial machine learning.
\end{IEEEbiography}

\begin{IEEEbiography}[{\includegraphics[width=0.8in,height=1in,clip,keepaspectratio]{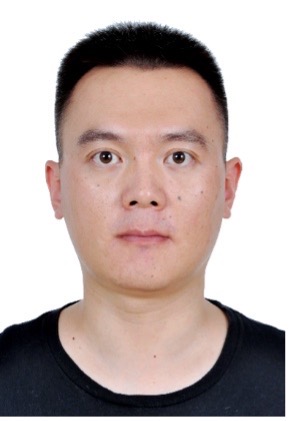}}]{Hua Zhang} is a professor with the Institute of Information Engineering, Chinese Academy of Sciences. He received the Ph.D. degree in computer science from the School of Computer Science and Technology, Tianjin University, Tianjin, China in 2015. His research interests include computer vision, multimedia, and machine learning.
\end{IEEEbiography}

\begin{IEEEbiography}
[{\includegraphics[width=0.8in,height=1in,clip,keepaspectratio]{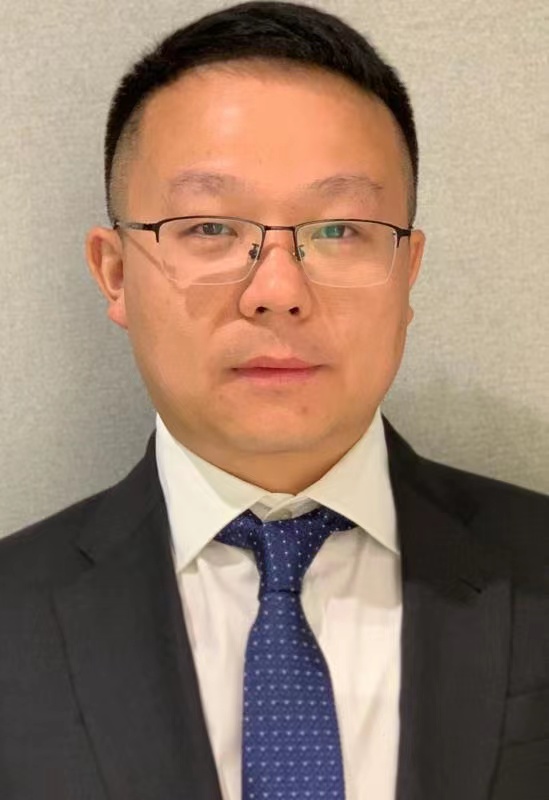}}]
{Xiaochun Cao}(SM'14)
received the B.S. and M.S. degrees in computer science from Beihang University, Beijing, China, and the Ph.D. degree in computer science from the University of Central Florida, Orlando, FL, USA. After graduation, he spent about three years at ObjectVideo Inc. as a Research Scientist. He is with the School of Cyber Science and Technology, Shenzhen Campus, Sun Yat-sen University, Shenzhen 518107, P.R. China. He has authored and co-authored more than 100 journal and conference papers.
Prof. Cao is a Fellow of the IET. He is on the Editorial Boards of the IEEE Transactions on Image Processing, IEEE Transactions on Multimedia, IEEE Transactions on Circuits and Systems for Video Technology. His dissertation was nominated for the University of Central Florida's university-level Outstanding Dissertation Award. In 2004 and 2010, he was the recipient of the Piero Zamperoni Best Student Paper Award at the International Conference on Pattern Recognition.
\end{IEEEbiography}
% \clearpage
% \appendix

% \input{tex/revised_appendix}

\end{document}